\documentclass[letterpaper, 10 pt, journal, twoside]{IEEEtran}   % Comment this line out if you need a4paper

\IEEEoverridecommandlockouts                              % This command is only needed if 
\usepackage[numbers,sort&compress]{natbib}
\usepackage{graphicx}
\usepackage{epsfig}
\usepackage{verbatim}
\usepackage{color}
\usepackage{soul}
\usepackage[font=footnotesize]{caption}
\usepackage{amssymb}
\usepackage{amsmath}
\usepackage{algorithm}
\usepackage{algpseudocode}
\usepackage[english]{babel}
\usepackage[T1]{fontenc}
\usepackage[utf8]{inputenc}
\usepackage{xspace}
\usepackage{todonotes}
\usepackage{epstopdf}
\usepackage{hyperref}
\usepackage{url}
\usepackage{physics}
\usepackage[nolist,nohyperlinks]{acronym}
\usepackage{tikz}
\usepackage{pgfplots}
\usepackage{subcaption}
\usepackage{makecell}
\usepackage{changes}
\usepackage[export]{adjustbox}
\usepackage{pdfpages}
\usetikzlibrary{shapes,arrows,positioning}
\usetikzlibrary{calc}
\usetikzlibrary{external}
\tikzstyle{block} = [draw, fill=blue!20, rectangle, 
    minimum height=3em, minimum width=6em]
\tikzstyle{sum} = [draw, fill=blue!20, circle, node distance=1cm]
\tikzstyle{input} = [coordinate]
\tikzstyle{output} = [coordinate]
\tikzstyle{pinstyle} = [pin edge={to-,thin,black}]

\title{%\LARGE \bf
X-View: Graph-Based Semantic Multi-View Localization
}

\author{Abel Gawel$^{*}$,
Carlo Del Don$^{*}$,
Roland Siegwart,
Juan Nieto and
Cesar Cadena%
\thanks{This work was supported by European Union's Seventh Framework Programme for research, technological development and demonstration under the TRADR project No. FP7-ICT-609763 and by the National Center of Competence in Research (NCCR) Robotics through the Swiss National Science Foundation.} %Use only for final RAL version
\thanks{$^{*}$ The authors contributed equally to this work.}%
\thanks{Authors are with the Autonomous Systems Lab, ETH Zurich. \tt\small gawela@ethz.ch, \tt\small deldonc@student.ethz.ch, \tt\small \{rsiegwart, nietoj, cesarc\}@ethz.ch}
}%

\begin{document}
\onecolumn
{\large
%\documentclass[letterpaper, 12 pt]{article}
%\setlength{\parindent}{0pt}
%\begin{document}

\thispagestyle{empty}

{\setlength{\parindent}{0cm}
\textcopyright 2018 IEEE. Personal use of this material is permitted. Permission from IEEE must be obtained
for all other uses, in any current or future media, including reprinting/republishing this material
for advertising or promotional purposes, creating new collective works, for resale or redistribution
to servers or lists, or reuse of any copyrighted component of this work in other works.
}
\\

{\setlength{\parindent}{0cm}
Pre-print of article that will appear in the 2018 IEEE Robotics and Automation Letters (RA-L).
}
\\

{\setlength{\parindent}{0cm}
Please cite this paper as:
}
\\

{\setlength{\parindent}{0cm}
A. Gawel, C. Del Don, R. Siegwart, J. Nieto, and C. Cadena. (2018).\\ "X-View: Graph-Based Semantic Multi-View Localization" in IEEE Robotics and Automation Letters (RA-L), 2018.
}
\\

{\setlength{\parindent}{0cm}
bibtex:
}
\\

{\setlength{\parindent}{0cm}
@inproceedings\{gawel2018x-view,\\
  \phantom{x} title=\{X-View: Graph-Based Semantic Multi-View Localization\},\\ 
  \phantom{x} author=\{Gawel, Abel and Del Don, Carlo and Siegwart, Roland and Nieto, Juan and Cadena, Cesar\},\\ 
  \phantom{x} booktitle={IEEE Robotics and Automation Letters (RA-L)\},\\
  \phantom{x} year=\{2018\}
  \\ 
\}
}

%\end{document}
\par}
\normalsize
\twocolumn
\clearpage
\setcounter{page}{1}
\maketitle

%%%%%%%%%%%%%%%%%%%%%%%%%%%%%%%%%%%%%%%%%%%%%%%%%%%%%%%%%%%%%%%%%%%%%%%%%%%%%%%%
\begin{abstract}
Global registration of multi-view robot data is a challenging task. 
Appearance-based global localization approaches often fail under drastic view-point changes, as representations have limited view-point invariance.
This work is based on the idea that human-made environments contain rich semantics which can be used to disambiguate global localization.
Here, we present \emph{X-View}, a Multi-View Semantic Global Localization system.
\emph{X-View} leverages semantic graph descriptor matching for global localization, enabling localization under drastically different view-points.
While the approach is general in terms of the semantic input data, we present and evaluate an implementation on visual data.
We demonstrate the system in experiments on the publicly available \emph{SYNTHIA} dataset, on a realistic urban dataset recorded with a simulator, and on real-world StreetView data. 
Our findings show that \emph{X-View} is able to globally localize aerial-to-ground, and ground-to-ground robot data of drastically different view-points.
Our approach achieves an accuracy of up to $85\,\%$ on global localizations in the multi-view case, while the benchmarked baseline appearance-based methods reach up to $75\,\%$.

\end{abstract}


%%%%%%%%%%%%%%%%%%%%%%%%%%%%%%%%%%%%%%%%%%%%%%%%%%%%%%%%%%%%%%%%%%%%%%%%%%%%%%%%
\section{INTRODUCTION}

\IEEEPARstart{G}{lobal} localization between heterogeneous robots is a difficult problem for classic place-recognition approaches.
Visual appearance-based approaches such as \cite{cummins2008fab, galvez2012bags} are currently among the most effective methods for re-localization.
However, they tend to significantly degrade with appearance changes due to different time, weather, season, and also view-point~\cite{lowry2016visual, arandjelovic2016netvlad}.
In addition, when using different sensor modalities, the key-point extraction becomes an issue as they are generated from different physical and geometrical properties, for instance intensity gradients in images vs. high-curvature regions in point clouds.

Relying on geometrical information, directly from the measurements or from a reconstruction algorithm, on the other hand shows stronger robustness on view-point changes, seasonal changes, and different sensor modalities.
However, geometrical approaches typically do not scale well to very large environments, and it remains questionable if very strong view-point changes can be compensated while maintaining only a limited overlap between the localization query and database~\cite{gawel2016structure, gawel20173d}.

Another avenue to address appearance and view-point changes are \ac{CNN} architectures for place recognition~\cite{arandjelovic2016netvlad, chen2017deep}.
While these methods show strong performance under appearance changes, their performance is still to be investigated under extreme view-point variations.

Recently, topological approaches to global localization regained interest as a way to efficiently encode relations between multiple local visual features~\cite{stumm2016robust, su2016fast}.
On the other hand, the computer vision community has made great progress in semantic segmentation and classification, resulting in capable tools for extracting semantics from visual and depth data~\cite{garcia2017review, valada2017adapnet, badrinarayanan2015segnet}.

\begin{figure}[t]
\includegraphics[width = 0.9\columnwidth]{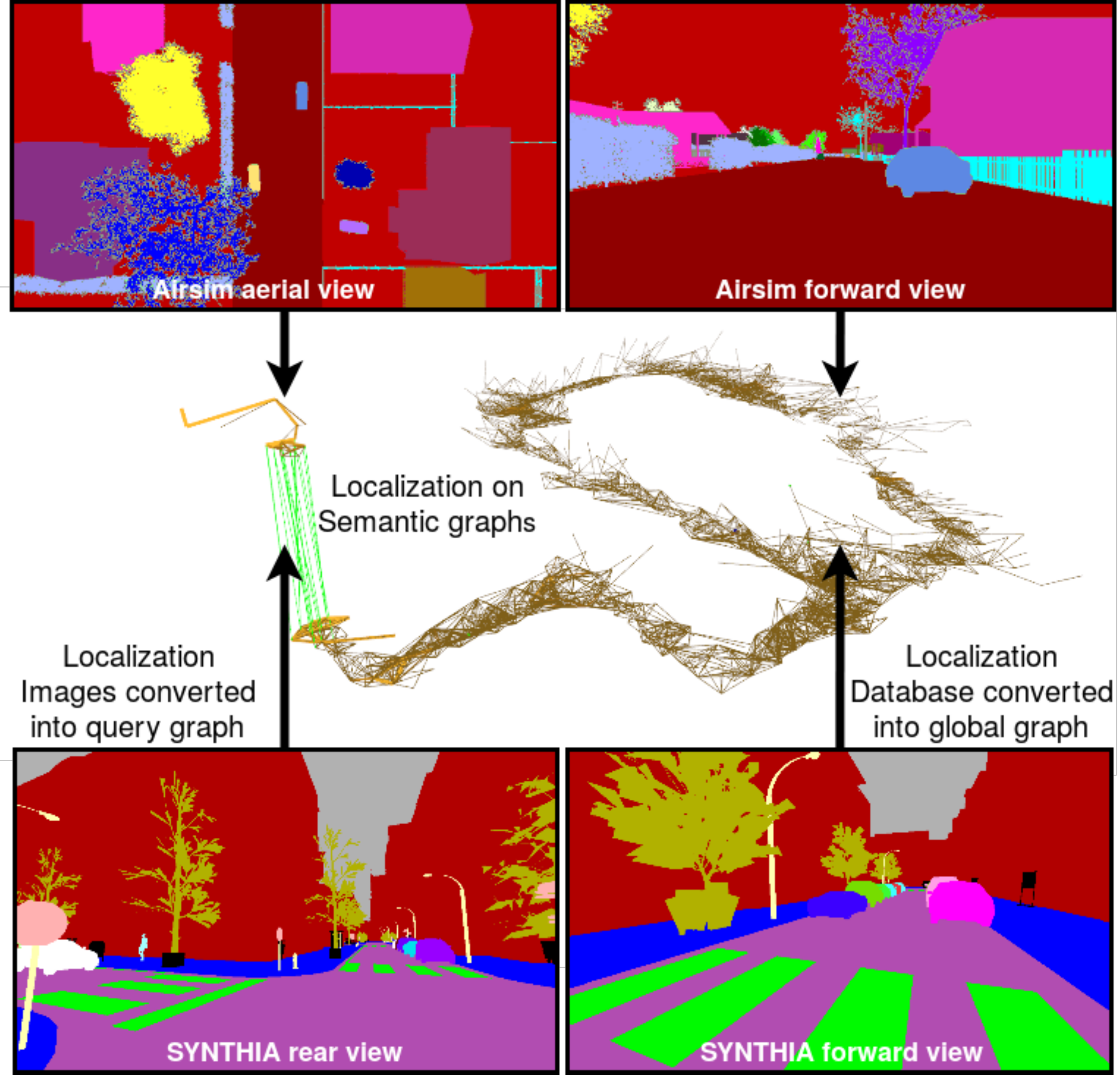}
\centering
\caption{\emph{X-View} globally localizes data of drastically different view-points using graph representations of semantic information. Here, samples of the experimental data is shown, i.e., semantically segmented images from the publicly available \emph{SYNTHIA} and the \emph{Airsim} datasets. The localization target graph is built from data of one view-point (\emph{right images}), while the query graph is built from sequences of another view-point (\emph{left images}). \emph{X-View} efficiently localizes the query graph in the target graph.}
\vspace{-1.5em}
\label{fig:datasets}
\end{figure}

Based on the hypothesis that semantics can help to mitigate the effects of appearance changes, we present \emph{X-View}, a novel approach for global localization based on building graphs of semantics.
\emph{X-View} introduces graph descriptors that efficiently represent unique topologies of semantic objects.
These can be matched in much lower computational effort, therefore not suffering under the need for exhaustive sub-graph matching \cite{cook1971complexity}.

By using semantics as an abstraction between robot view-points, we achieve invariances to strong view-point changes, outperforming \ac{CNN}-based techniques on RGB data.
Furthermore, with semantics understanding of the scene, unwanted elements, such as moving objects can naturally be excluded from the localization.
We evaluate our global localization algorithm on publicly available datasets of real and simulated urban outdoor environments, and report our findings on localizing under strong view-point changes.
Specifically, this paper presents the following contributions:
\begin{itemize}
\item A novel graph representation for semantic topologies. 
\item Introduction of a graph descriptor based on random walks that can be efficiently matched with established matching methods.
\item A full pipeline to process semantically segmented images into global localizations.
\item Open source implementation of the \emph{X-View} algorithm\footnote{\href{https://github.com/ethz-asl/x-view}{https://github.com/ethz-asl/x-view}}.
\item Experimental evaluation on publicly available datasets.
\end{itemize}
The remainder of this paper is structured as follows:
Sec.~\ref{sec:related_work} reviews the related work on global localization, followed by the presentation of the \emph{X-View} system in Sec.~\ref{sec:x_view}.
We present our experimental evaluation in Sec.~\ref{sec:experiments} and conclude our findings in Sec.~\ref{sec:conclusions}.
\section{RELATED WORK}
\label{sec:related_work}
In this section we review the current state-of-the-art in multi-robot global localization in relation to our proposed system.

A common approach to global localization is visual feature matching. 
A large amount of approaches have been proposed in the last decade, giving reliable performance under perceptually similar conditions~\cite{galvez2012bags, lowry2016visual, cummins2008fab}.
Several extensions have been proposed to overcome perceptually difficult situations, such as seasonal changes \cite{milford2012seqslam, cieslewski2016point}, daytime changes~\cite{burki2016appearance,arandjelovic2016netvlad}, or varying view-points using \ac{CNN} landmarks~\cite{sunderhauf2015place, chen2017deep}.
However, drastic view-point invariance, e.g., between views of aerial and ground robots continues to be a challenging problem for appearance-based techniques.

In our previous work, we demonstrated effective \emph{3D} heterogeneous map merging approaches between different view-points from camera and LiDAR data, based on overlapping \emph{3D} structural descriptors~\citep{gawel2016structure, gawel20173d}.
However, 3D reconstructions are still strongly view-point dependent.
While these techniques do not rely on specific semantic information of the scenes, the scaling to large environments has not yet been investigated, and computational time is outside real-time performance with large maps.

Other approaches to global localization are based on topological mapping \cite{huang2005topological, marinakis2010pure}.
Here, maps are represented as graphs $\boldsymbol{G} = (\boldsymbol{V}, \boldsymbol{E})$ of unique vertices $\boldsymbol{V}$ and edges $\boldsymbol{E}$ encoding relationships between vertices.
While these works focus on graph merging by exhaustive vertex matching on small graphs, they do not consider graph extraction from sensory data or ambiguous vertices.
Furthermore, the computationally expensive matching does not scale to larger graph comparisons.

With the recent advances in learning-based semantic extraction methods, using semantics for localization is a promising avenue \cite{kostavelis2015semantic, bowman2017probabilistic, atanasov2014semantic}.
In \cite{bowman2017probabilistic, atanasov2014semantic} the authors focus on the \emph{data association} problem for semantic localization using \ac{EM} and the formulation of the pose estimation problem for semantic constraints as an error minimization.
The semantic extraction is based on a standard object detector from visual key-points.

\citet{stumm2016robust} propose to use graph kernels for place recognition on visual key-point descriptors.
Graph kernels are used to project image-wise covisibility graphs into a feature space.
The authors show that graph descriptions can help localization performance as to efficiently cluster multiple descriptors meaningfully.
However, the use of large densely connected graphs sets limitations to the choice of graph representation. 
Motivated, by these findings, we propose to use graph descriptors on sparse semantic graphs for global localization.
\section{X-VIEW}
\label{sec:x_view}
In this section, we present our Graph-Based Multi-View Semantic Global Localization system, coined \emph{X-View}.
It leverages graph extraction from semantic input data and graph matching using graph descriptors.
Fig.~\ref{fig:overview} illustrates the architecture of the proposed global localization algorithm, focusing on the graph representation and matching of query semantic input data to a global graph.
The localization target map is represented as the global graph.
\emph{X-View} is designed to operate on any given odometry estimation system and semantic input cue.
However, for the sake of clarity, we present our system as implemented for semantically segmented images, but it is not limited to it.
\pgfdeclarelayer{background}
\pgfdeclarelayer{foreground}
\pgfsetlayers{background,main,foreground}
\begin{figure*}
\centering
\includegraphics[width = 0.95\textwidth]{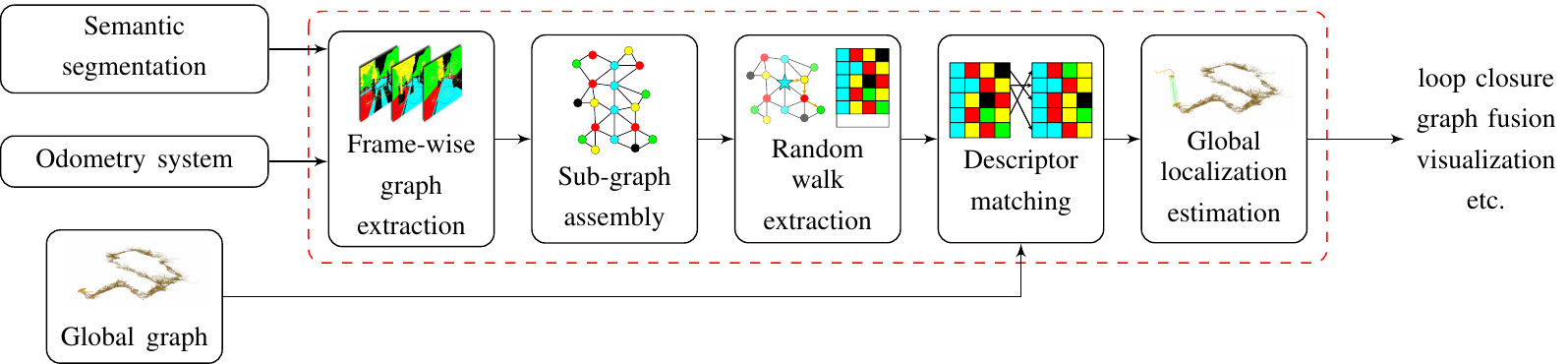}
\caption{\emph{X-View} global localization system overview.
The inputs to the system are semantically segmented frames (e.g., from RGB images) and the global graph $\boldsymbol{G}_{db}$. 
First, a local graph is extracted from the new segmentation. 
Then, the sub-graph $\boldsymbol{G}_{q}$ is assembled and random walk descriptors are computed on each node of $\boldsymbol{G}_{q}$. The system matches the sub-graph random walk descriptors to $\boldsymbol{G}_{db}$, e.g., recorded from a different view-point. 
Finally, the matches are transferred to the localization back-end module to estimate the relative localization between $\boldsymbol{G}_{q}$ and $\boldsymbol{G}_{db}$.
Consecutively, the relative localization can be used for various purposes such as loop closure, fusing $\boldsymbol{G}_{q}$ into $\boldsymbol{G}_{db}$ or for visualization.}
\vspace{-1.5em}
\label{fig:overview}
\end{figure*}
\subsection{System input}
We use semantically segmented images containing pixel-wise semantic classification as input to the localization algorithm.
These segmentations can be achieved using a semantic segmentation method, such as \cite{badrinarayanan2015segnet, valada2017adapnet}.
Also instance-wise segmentation, i.e., unique identifiers for separating overlapping objects of same class in the image space can be considered for improved segmentation, but is not strictly necessary for the approach to work.
Furthermore, we assume the estimate of an external odometry system.
Finally, we also consider a database semantic graph $\boldsymbol{G}_{db}$, as it could have been built and described on a previous run of our graph building algorithm as presented in the next sub-sections.
\subsection{Graph extraction and assembly}
\label{sec:extraction}

In this step, we convert a sequence of semantic images $\boldsymbol{I}_q$ into a query graph $\boldsymbol{G}_q$.
We extract blobs of connected regions, i.e., regions of the same class label $l_j$ in each image.
Since semantically segmented images often show noisy partitioning of the observed scene (holes, disconnected edges and invalid labels on edges), we smooth them by dilating and eroding the boundaries of each blob.
We furthermore reject blobs smaller than a minimum pixel count to be included in the graph, to mitigate the effect of minor segments.
This process removes unwanted noise in the semantically segmented images.
The magnitude of this operation is $4$ pixels, and has a minor effect on the segmentation result.
However, it ensures clean boundaries between semantic segments.
Furthermore, the center location $\boldsymbol{p}_j$ of the blobs are extracted and stored alongside the blob labels as vertices $\boldsymbol{\mathbf{v}}_j = \{\boldsymbol{l}_j, \boldsymbol{p}_j\}$.
In the case that also instance-wise segmentation is available, it can be considered in the blob extraction step, otherwise the extraction operates only on a class basis.

The undirected edges $\boldsymbol{e}_j$ between vertices are formed when fulfilling a proximity requirement, which can be either in image- or \emph{3D}-space.
In the case of image-space, we assume images to be in a temporal sequence to grow graphs over several frames of input data.
However, this is not required in the \emph{3D} case.

Using a depth channel or the depth estimation from, e.g., optical flow, the neighborhood can be formed in \emph{3D}-space, using the \emph{3D} locations of the image blobs to compute a Euclidean distance.
The process is illustrated for image data in Fig.~\ref{fig:graph_sequence} (top).
\begin{figure}
\includegraphics[width = 0.9\columnwidth]{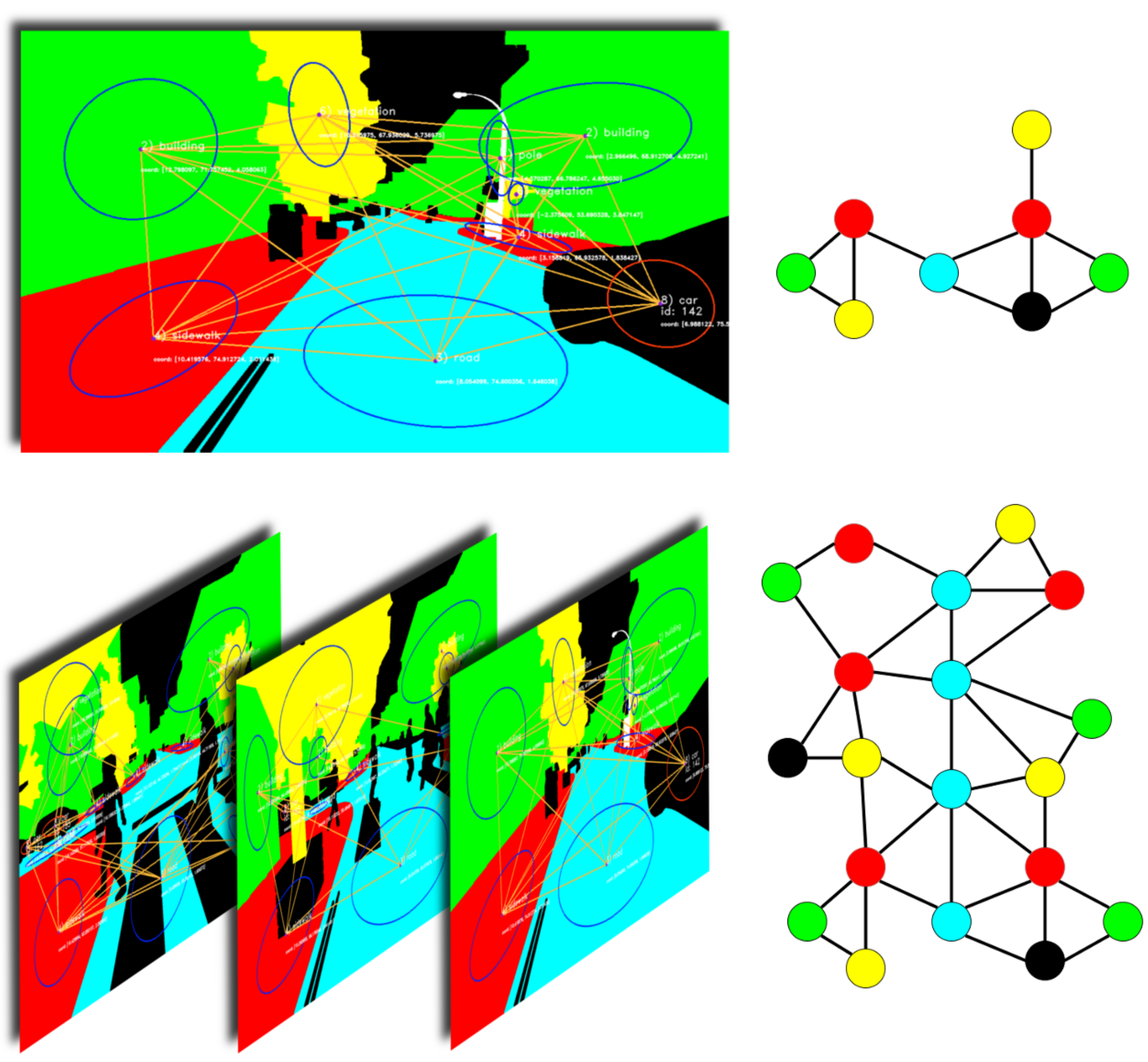}
\centering
\caption{Extraction of semantic graphs from one image (top) and a sequence of images (bottom). 
Vertices are merged and connected from sequences of input data. 
Note that we omitted some vertices and edges in the sample graphs on the right side for visualization purposes and reduced the graph to a planar visualization, whereas the semantic graphs in our system are connected in \emph{3D}-space. The ellipses around each vertex were added for visualization and represent a scaled fitted ellipse on a semantic instance of the segmentation image.
}
\vspace{-1.5em}
\label{fig:graph_sequence}
\end{figure}
Then, several image-wise graphs are merged into $\boldsymbol{G}_q$ by connecting vertices of consecutive images using their Euclidean distance, see Fig.~\ref{fig:graph_sequence}.
To prevent duplicate vertices of the same semantic instance, close instances in $\boldsymbol{G}_q$ are merged into a single vertex, at the location of the vertices' first observation.
The strategy of merging vertices into their first observation location is further motivated by the structure of \emph{continuous} semantic entities, such as streets.
This strategy leads to evenly spaced creation of \emph{continuous} entities' vertices in $\boldsymbol{G}_q$.
\subsection{Descriptors}
\label{sec:description}
\emph{X-View} is based on the idea that semantic graphs hold high descriptive power, and that localizing a sub-graph in a database graph can yield good localization results.
However, since sub-graph matching is an NP-complete problem \cite{cook1971complexity}, a different regime is required to perform the graph registration under real-time constraints, i.e., in the order of seconds for typical robotic applications.
In this work, we extract random walk descriptors for every node of the graph \cite{perozzi2014deepwalk}, and match them in a subsequent step.
This has the advantage that the descriptors can be extracted and matched in constant or linear time, given a static or growing database-graph, respectively.

Each vertex descriptor is an $n\times m$ matrix consisting of $n$ random walks of depth $m$.
Each of the random walks originates at the base vertex $\boldsymbol{\mathbf{v}}_j$ and stores the class labels of the visited vertices.
Walk strategies, such as preventing from immediate returns to the vertex that was visited in the last step, and exclusion of duplicate random walks can be applied to facilitate expressiveness of the random walk descriptors.
The process of random walk descriptor extraction is illustrated in Fig.~\ref{fig:random_walk}.
\begin{figure}
\includegraphics[width = 0.7\columnwidth]{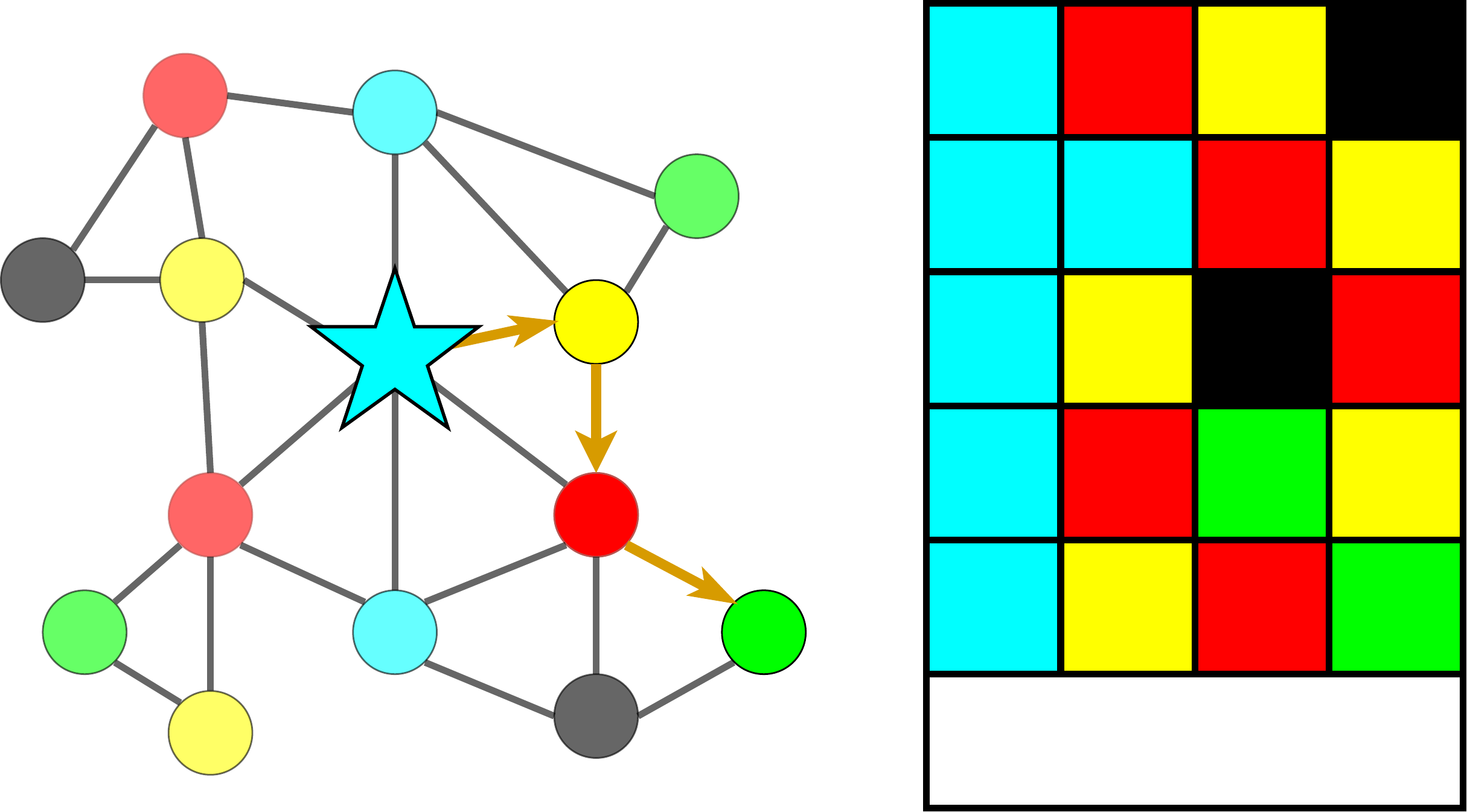}
\centering
\caption{Schematic representation of the random walk extraction.
(Left) From a seed vertex, cyan star, the random walker explores its neighborhood. 
This results in the descriptor of $n$ random walks of depth $m$ (here, $m=4$).
The highlighted path corresponds to the last line of the descriptor on the right.
(Right) Each line of the descriptor starts with the seed vertex label and continues with the class labels of the visited vertices.}
\vspace{-1.5em}
\label{fig:random_walk}
\end{figure}
\subsection{Descriptor Matching}
\label{sec:matching}
After both $\boldsymbol{G}_q$ and $\boldsymbol{G}_{db}$ are created, we find associations between vertices in the query graph and the ones in the database graph by computing a similarity score between the corresponding graph descriptors.
The similarity measure is computed by matching each row of the semantic descriptor of the query vertex to the descriptor of the database vertex.
The number of identical random walks on the two descriptors reflects the similarity score $s$, which is normalized between $0$ and $1$.
In a second step, the $k$ matches with highest similarity score are selected for estimating the location of the query graph inside the database map.
\subsection{Localization Back-End}

The matching between query graph and global graph, the robot-to-vertex observations, and the robot odometry measurements result in constraints $\boldsymbol{\theta}_i \subseteq \boldsymbol{\Theta}(\boldsymbol{p}_{i}, \boldsymbol{c}_{i})$ on the vertex positions $\boldsymbol{p}_{i}$ and robot poses $\boldsymbol{c}_{i}$ with $ \boldsymbol{\theta}_i= \boldsymbol{e}_{i}^T \boldsymbol{\Omega}_{i}\boldsymbol{e}_{i}$, the measurement errors $\boldsymbol{e}_{i}$, and associated information matrix $\boldsymbol{\Omega}_{i}$.
Specifically these three types of constraints are denoted as $\boldsymbol{\Theta}_{\mathbf{M}}(\boldsymbol{p}_i), \boldsymbol{\Theta}_{\mathbf{V}}(\boldsymbol{p}_i,\boldsymbol{c}_i)$, and $\boldsymbol{\Theta}_{O}(\boldsymbol{c}_i)$ respectively.
The matching constraints $\boldsymbol{\Theta}_{\mathbf{M}}(\boldsymbol{p}_i)$ stem from the semantic descriptor matching of the previous step, while the robot odometry constraints $\boldsymbol{\Theta}_{O}(\boldsymbol{c}_i)$ are created using the robots estimated odometry between consecutive robot poses associated to the localization graph.
The robot-to-vertex constraints encode the transformation between each robot-to-vertex observation.
Using these constraints, we compute a \ac{MAP} estimate of the robot pose $\boldsymbol{c}_i$ by minimizing a negative log-posterior $\mathbf{E} = \sum{\boldsymbol{\theta}_i}$, i.e.,
\begin{equation}
\boldsymbol{c}_i^* = \underset{\boldsymbol{c}_i}{\operatorname{argmin}} \sum \boldsymbol{\Theta}(\boldsymbol{p}_i,\boldsymbol{c}_{i})
\end{equation}
with $\boldsymbol{\Theta}(\boldsymbol{p}_{i}, \boldsymbol{c}_{i}) = \{\boldsymbol{\Theta}_M(\boldsymbol{p}_{i}), \boldsymbol{\Theta}_V(\boldsymbol{p}_{i}, \boldsymbol{c}_i), \boldsymbol{\Theta}_V(\boldsymbol{p}_{i})\}$
This optimization is carried out by a non-linear Gauss-Newton optimizer.
Optionally, the algorithm also allows to reject matching constraints in a sample consensus manner, using RANSAC on all constraints between $\boldsymbol{G}_q$ and $\boldsymbol{G}_{db}$, excluding the specific constraints from the optimization objective.
We initialize the robot position at the mean location of all matching vertices' locations from $\boldsymbol{G}_{db}$.
\section{EXPERIMENTS}
\label{sec:experiments}
We evaluate our approach on two different synthetic outdoor datasets with forward to rear view, and forward to aerial view, and one real world outdoor dataset with forward to rear view.
In this section, we present the experimental set-up, the results, and a discussion.
\subsection{Datasets}
The first of the used datasets is the public \emph{SYNTHIA} dataset~\cite{ros2016synthia}. 
It consists of several sequences of simulated sensor data from a car travelling in different dynamic environments and under varying conditions, e.g., weather and daytime.
The sensor data provides RGB, depth and pixel-wise semantic classification for 8 cameras, with always 2 cameras facing forward, left, backwards and right respectively.
The segmentation provides 13 different semantic classes which are labelled class-wise.
Additionally, dynamic objects, such as pedestrians and cars are also labelled instance-wise.
We use sequence 4, which features a town-like environment.
The total travelled distance is $970\,m$.

In the absence of suitable public aerial-ground semantic localization datasets, we use the photo-realistic \emph{Airsim} framework~\cite{airsim2017fsr} to generate a simulated rural environment\footnote{\href{http://robotics.ethz.ch/~asl-datasets/x-view/}{http://robotics.ethz.ch/\textasciitilde asl-datasets/x-view/}}.
This environment is explored with a top-down viewing \ac{UAV} and a car traversing the streets with forward-facing sensors.
Both views provide RGB, depth and pixel-wise semantic classification data in 13 different classes with instance-wise labelling.
Furthermore, both trajectories are overlapping with only an offset in $\mathbf{z}$-direction and have a length of $500\,m$ each.
Please note that we used a pre-built environment, i.e., the objects in the environment have not specifically been placed for enhanced performance.

Finally, we evaluate the system on a dataset gathered from \emph{Google StreetView} imagery.
The RGB and depth data of a straight $750\,m$ stretch  of Weinbergstrasse in Zurich are extracted via the \emph{Google Maps API}\footnote{\href{https://goo.gl/iBniJ9}{https://goo.gl/iBniJ9}}. 
Analogously to the \emph{SYNTHIA} dataset, we use forward and backward facing camera views.

While the travelled distance between two image locations in the \emph{Airsim} dataset is always $1\,m$, it varies between $0\,m$ to $1\,m$ in the \emph{SYNTHIA} dataset, and is approximately $10\,m$ between two frames in the \emph{StreetView} dataset.
Sample images of all datasets are depicted in Fig.~\ref{fig:mosaic}.
\begin{figure}
\begin{subfigure}{0.24\columnwidth}
\resizebox{\textwidth}{!}
{\includegraphics[trim={0 0.3cm 0 0.3cm},clip]{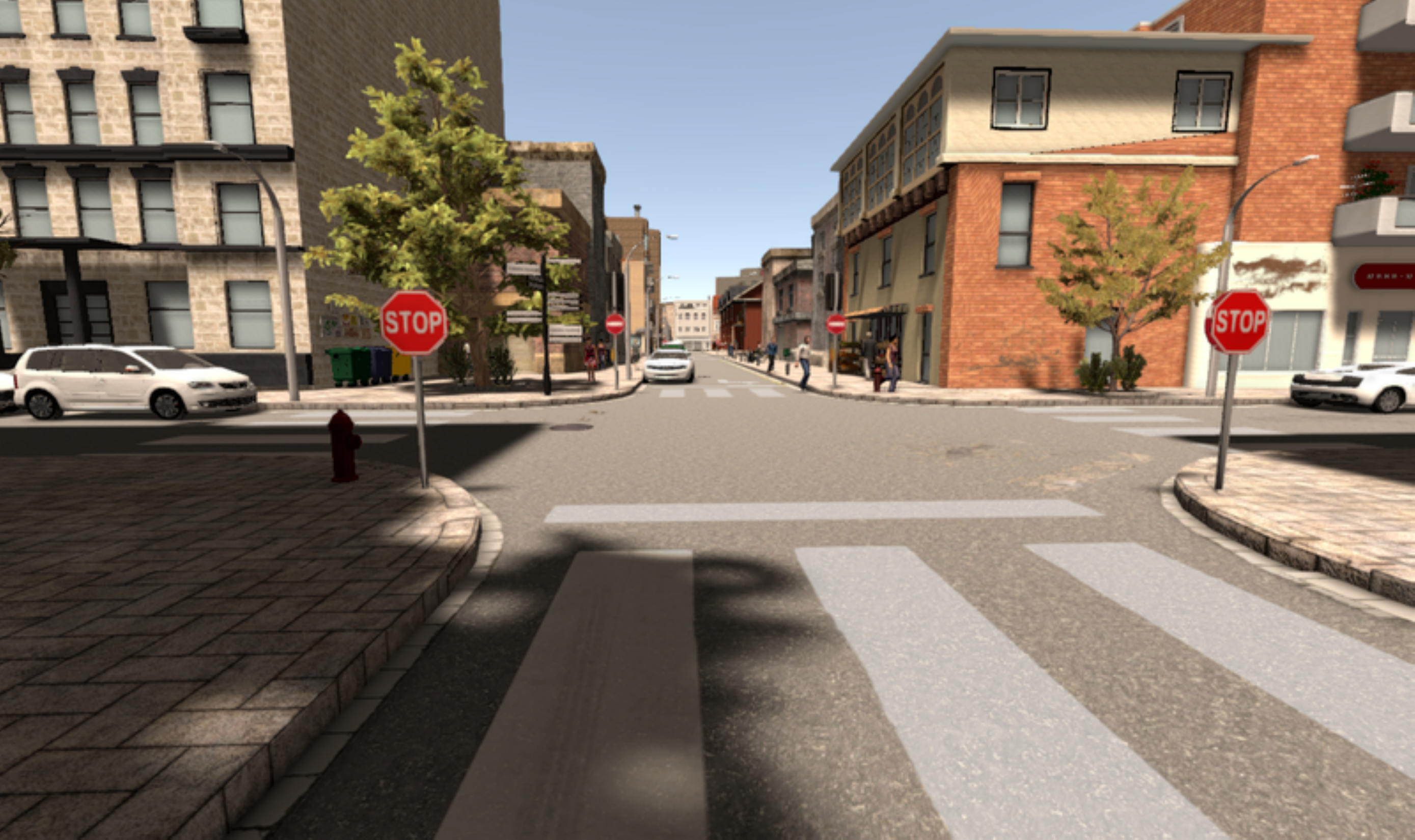}}
\label{fig:synthia_rgb}
\end{subfigure}\hspace*{\fill}
\begin{subfigure}{0.24\columnwidth}
\resizebox{\textwidth}{!}{\includegraphics[trim={0 0.3cm 0 0.3cm},clip]{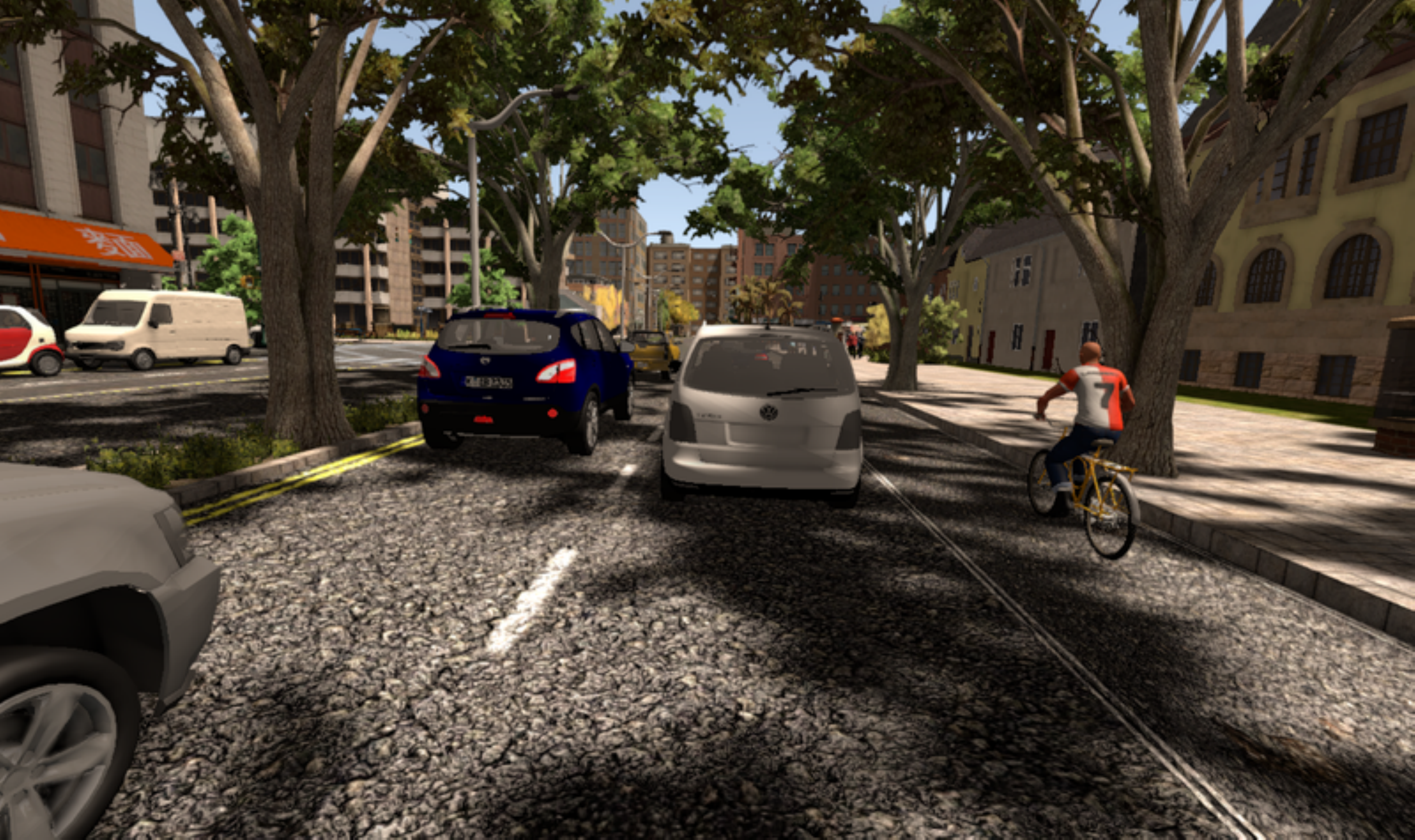}}
\label{fig:adapnet_rgb}
\end{subfigure}\hspace*{\fill}
\begin{subfigure}{0.24\columnwidth}
\resizebox{\textwidth}{!}{\includegraphics{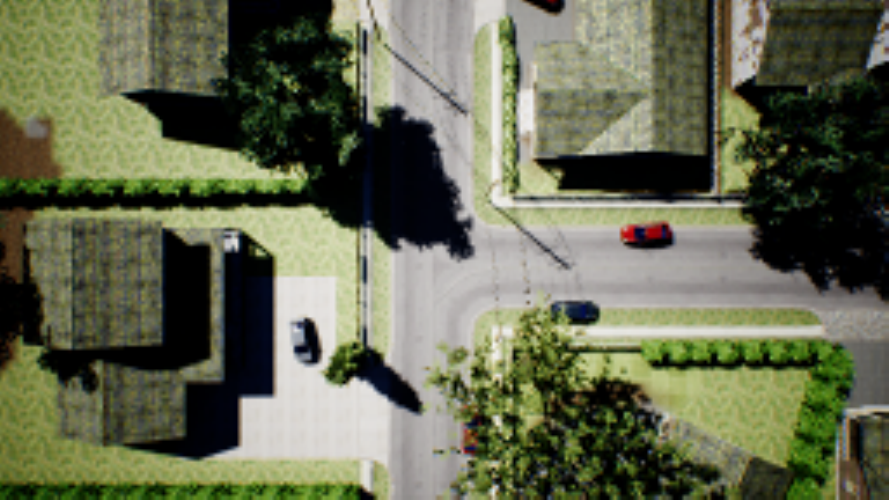}}
\label{fig:airsim_rgb}
\end{subfigure}\hspace*{\fill}
\begin{subfigure}{0.24\columnwidth}
\resizebox{\textwidth}{!}{\includegraphics[trim={0 2.2cm 0 2.2cm},clip]{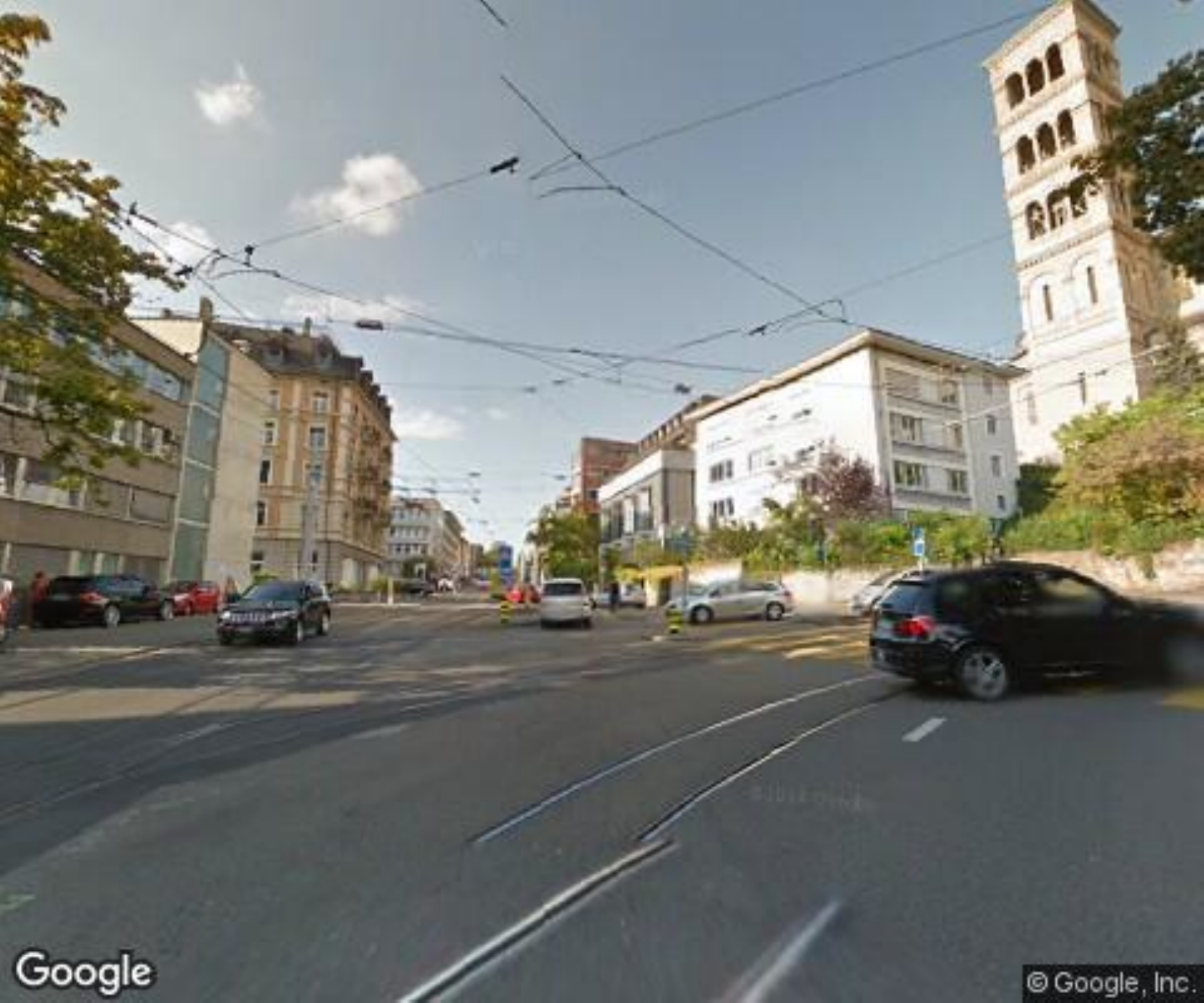}}
\label{fig:streetview_rgb}
\end{subfigure}\hspace*{\fill}
\smallbreak
\vspace{-1.4em}
\begin{subfigure}{0.24\columnwidth}
\resizebox{\textwidth}{!}{\includegraphics[trim={0 0.3cm 0 0.3cm},clip]{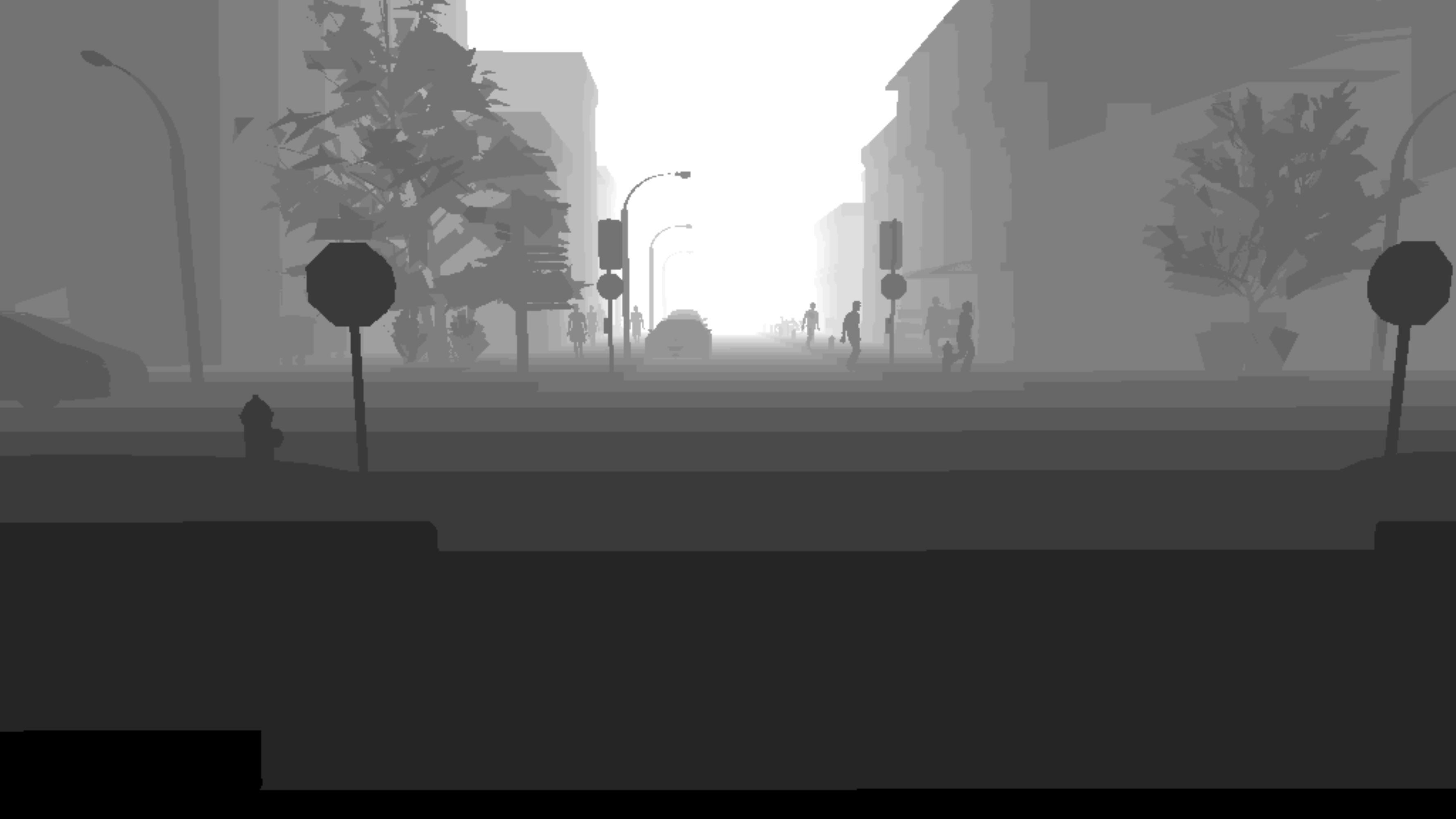}}
\label{fig:synthia_depth}
\end{subfigure}\hspace*{\fill}
\begin{subfigure}{0.24\columnwidth}
\resizebox{\textwidth}{!}{\includegraphics[trim={0 0.7cm 0 0.7cm},clip]{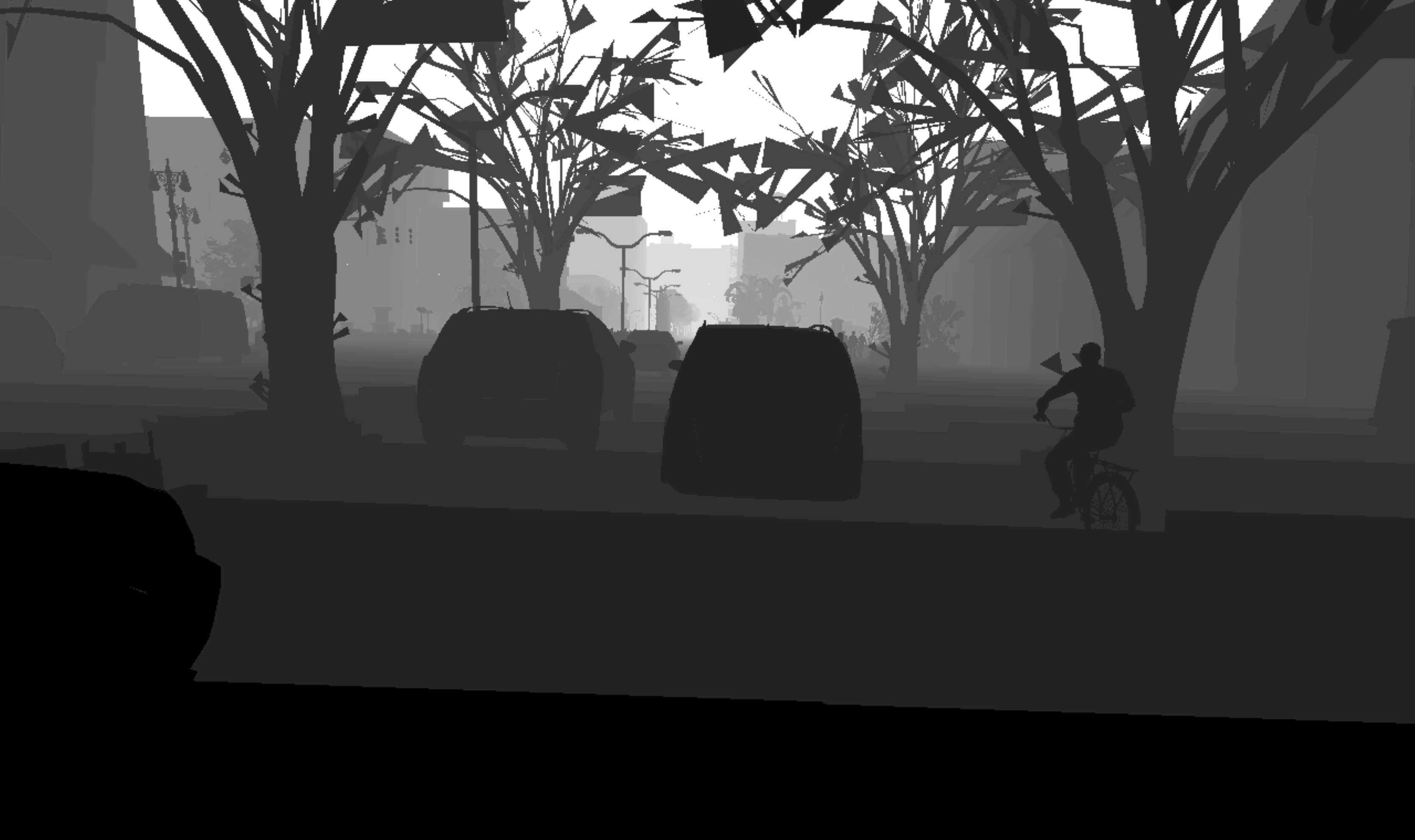}}
\label{fig:adapnet_depth}
\end{subfigure}\hspace*{\fill}
\begin{subfigure}{0.24\columnwidth}
\resizebox{\textwidth}{!}{\includegraphics{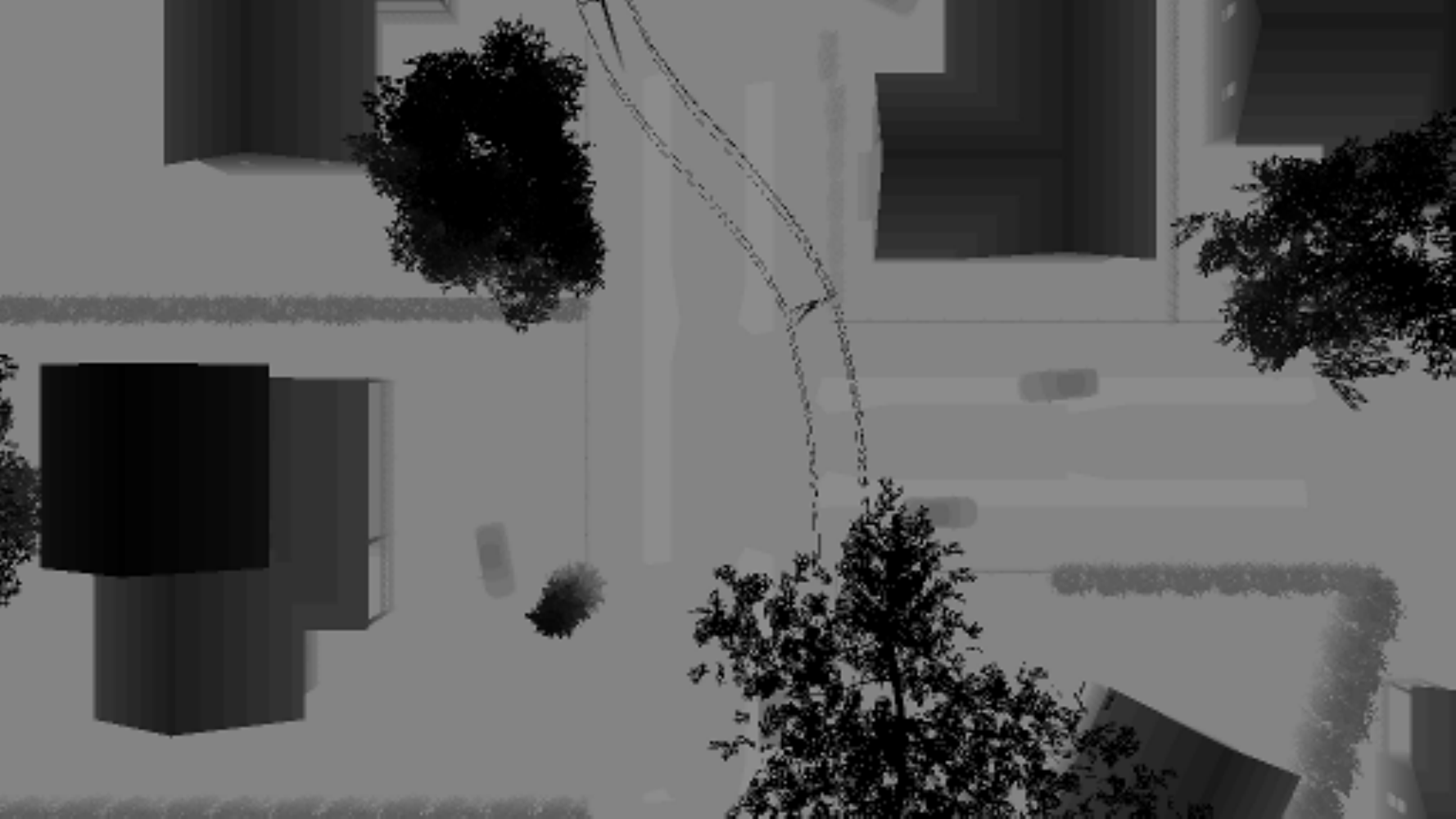}}
\label{fig:airsim_depth}
\end{subfigure}\hspace*{\fill}
\begin{subfigure}{0.24\columnwidth}
\resizebox{\textwidth}{!}{\includegraphics[trim={0 1.6cm 0 1.6cm},clip]{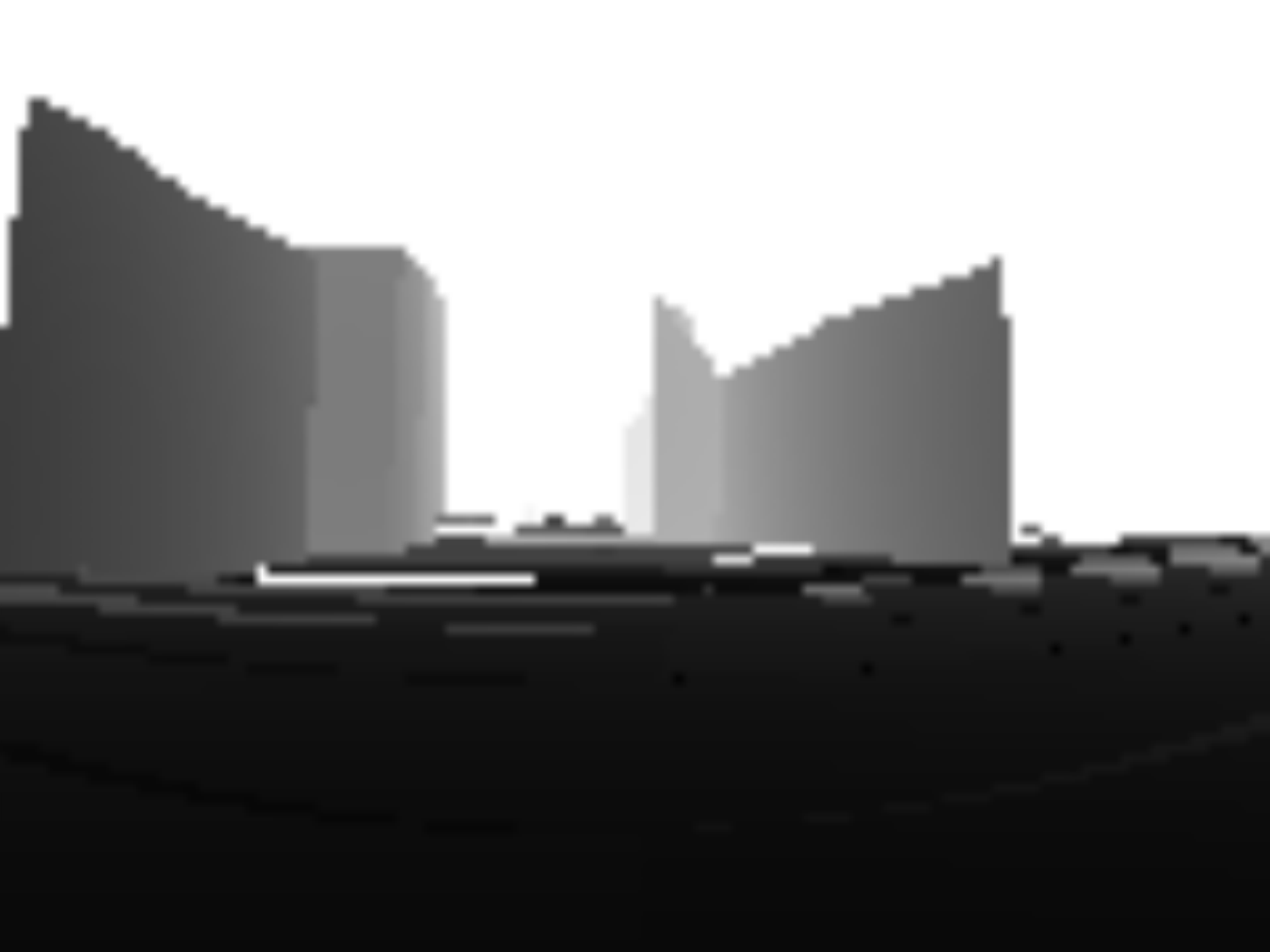}}
\label{fig:streetview_depth}
\end{subfigure}\hspace*{\fill}
\smallbreak
\vspace{-1.4em}
\begin{subfigure}{0.24\columnwidth}
\resizebox{\textwidth}{!}{\includegraphics[trim={0 0.5cm 0 0.5cm},clip]{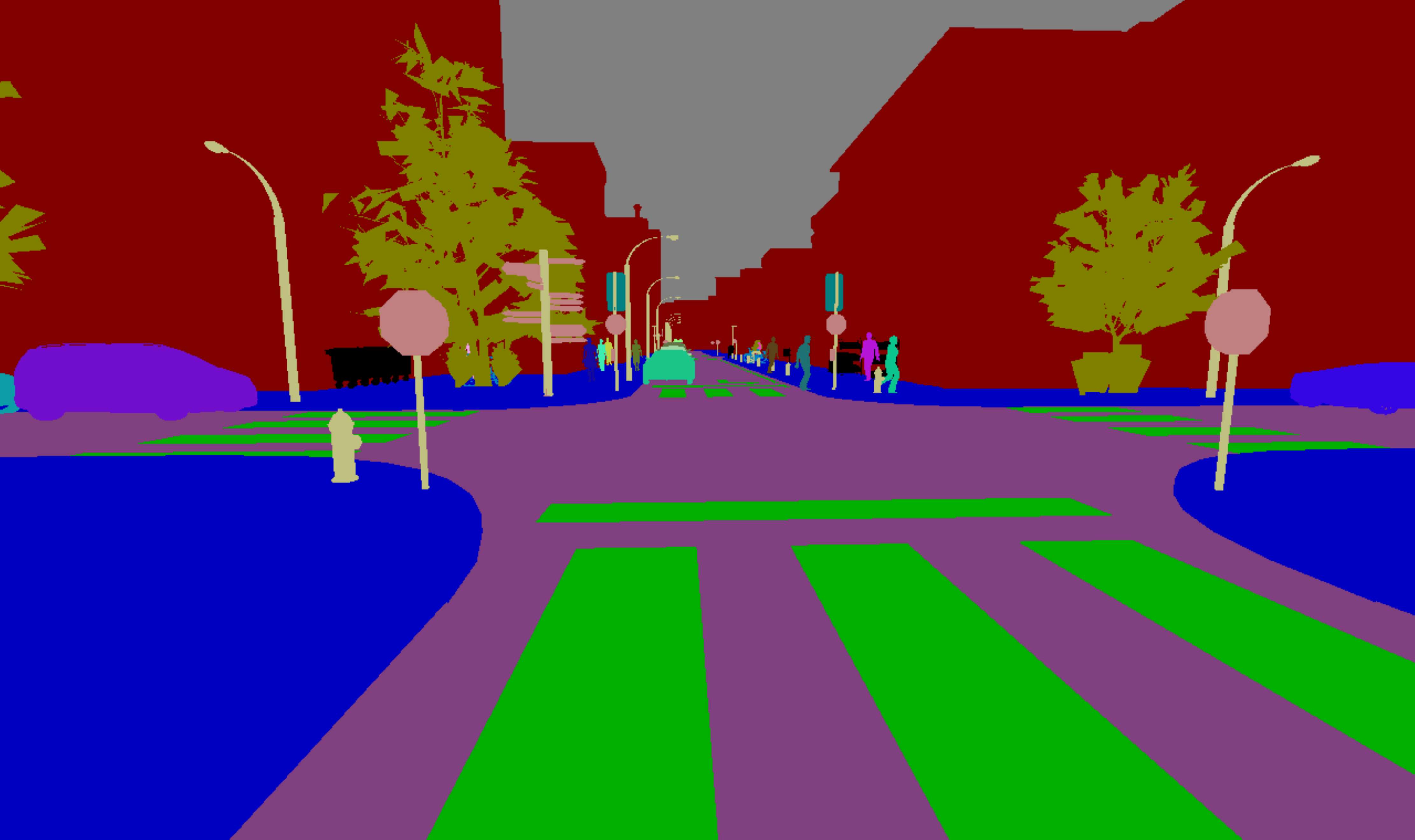}}
\label{fig:synthia_seg}
\vspace{-1.4em}
\caption*{SYNTHIA}
\end{subfigure}\hspace*{\fill}
\begin{subfigure}{0.24\columnwidth}
\resizebox{\textwidth}{!}{\includegraphics[trim={0 0.2cm 0 0.2cm},clip]{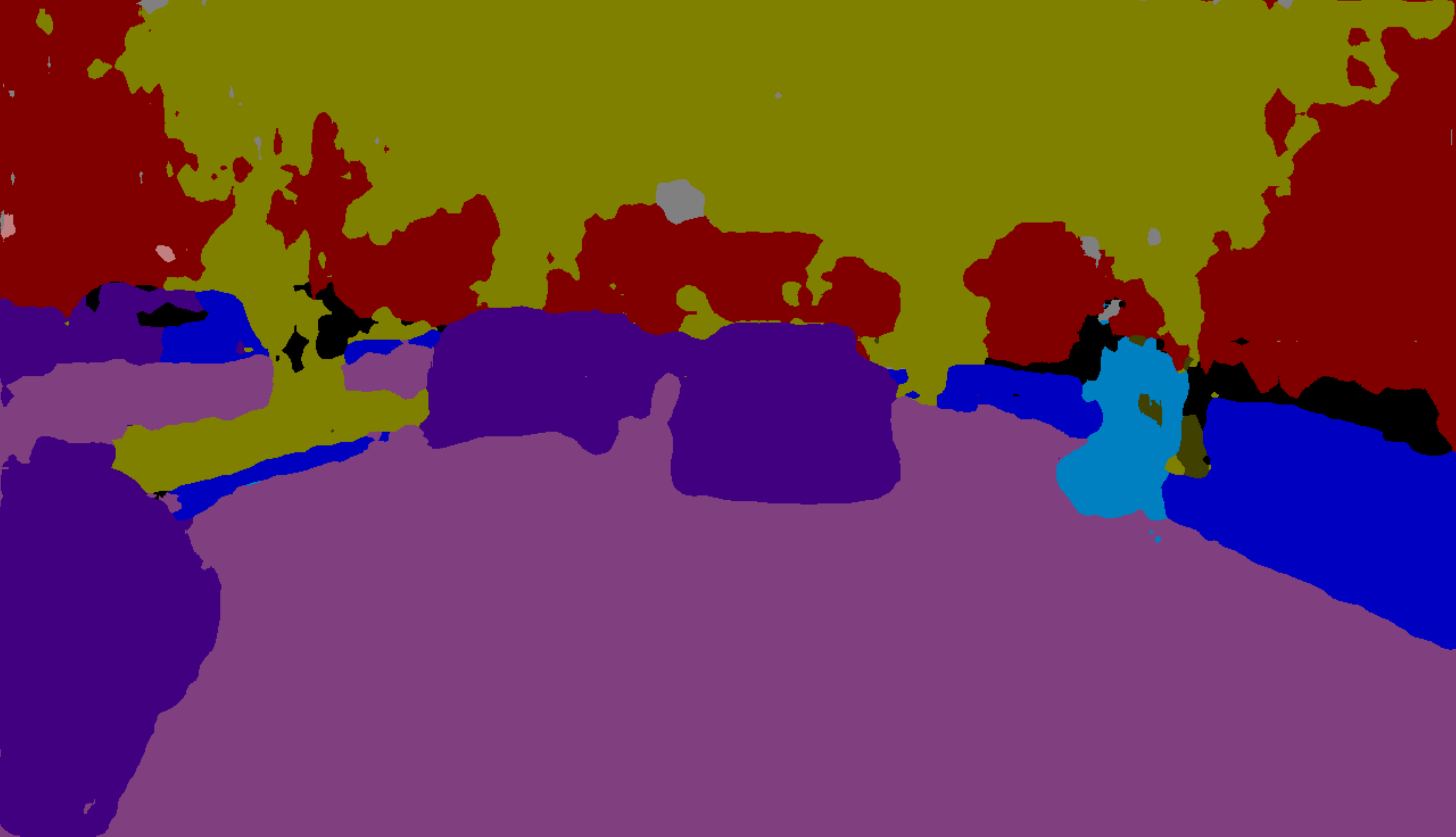}}
\label{fig:adapnet_seg}
\vspace{-1.4em}
\caption*{AdapNet}
\end{subfigure}\hspace*{\fill}
\begin{subfigure}{0.24\columnwidth}
\resizebox{\textwidth}{!}{\includegraphics{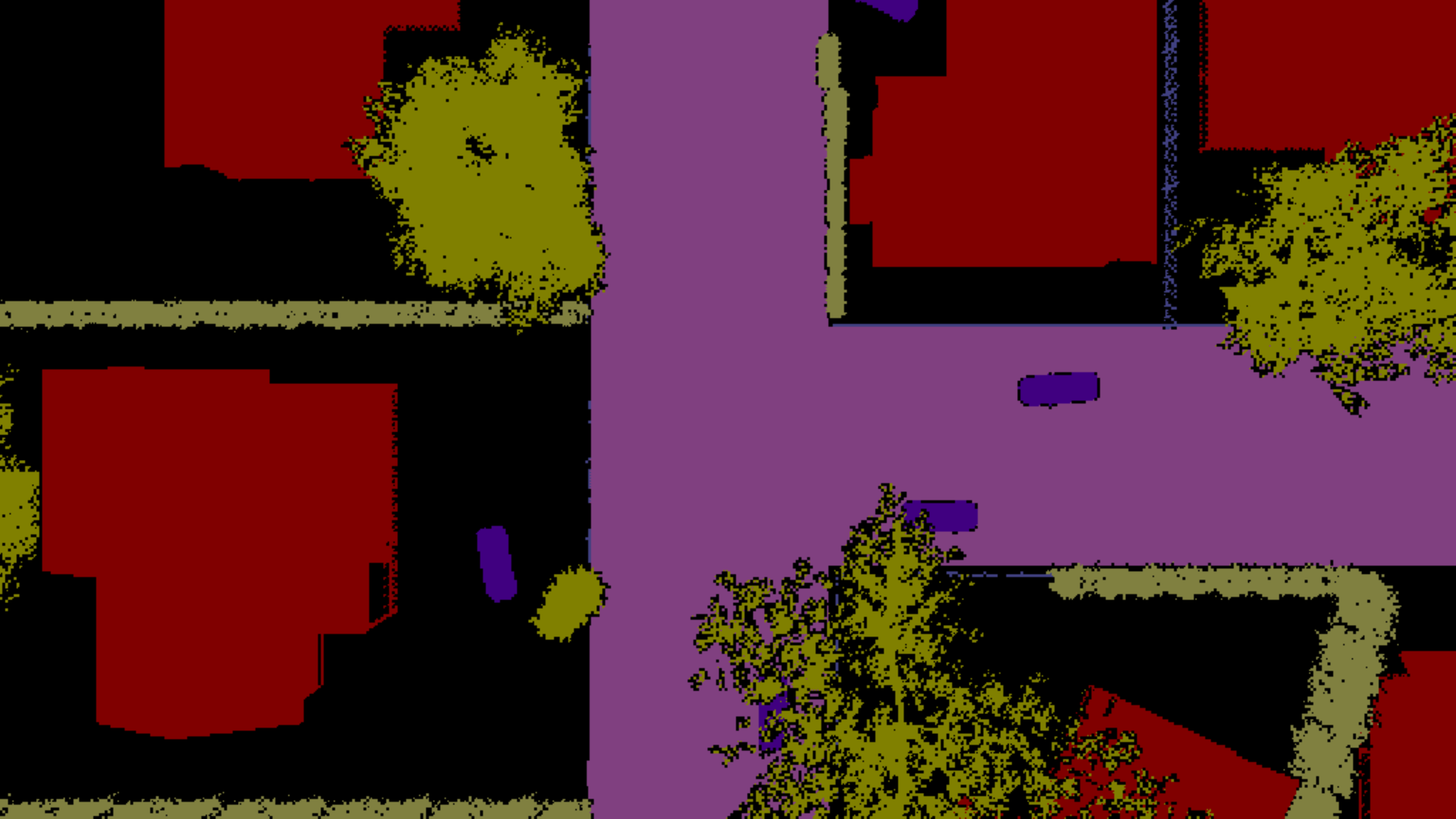}}
\label{fig:airsim_seg}
\vspace{-1.4em}
\caption*{Airsim}
\end{subfigure}\hspace*{\fill}
\begin{subfigure}{0.24\columnwidth}
\resizebox{\textwidth}{!}{\includegraphics[trim={0 1.6cm 0 1.6cm},clip]{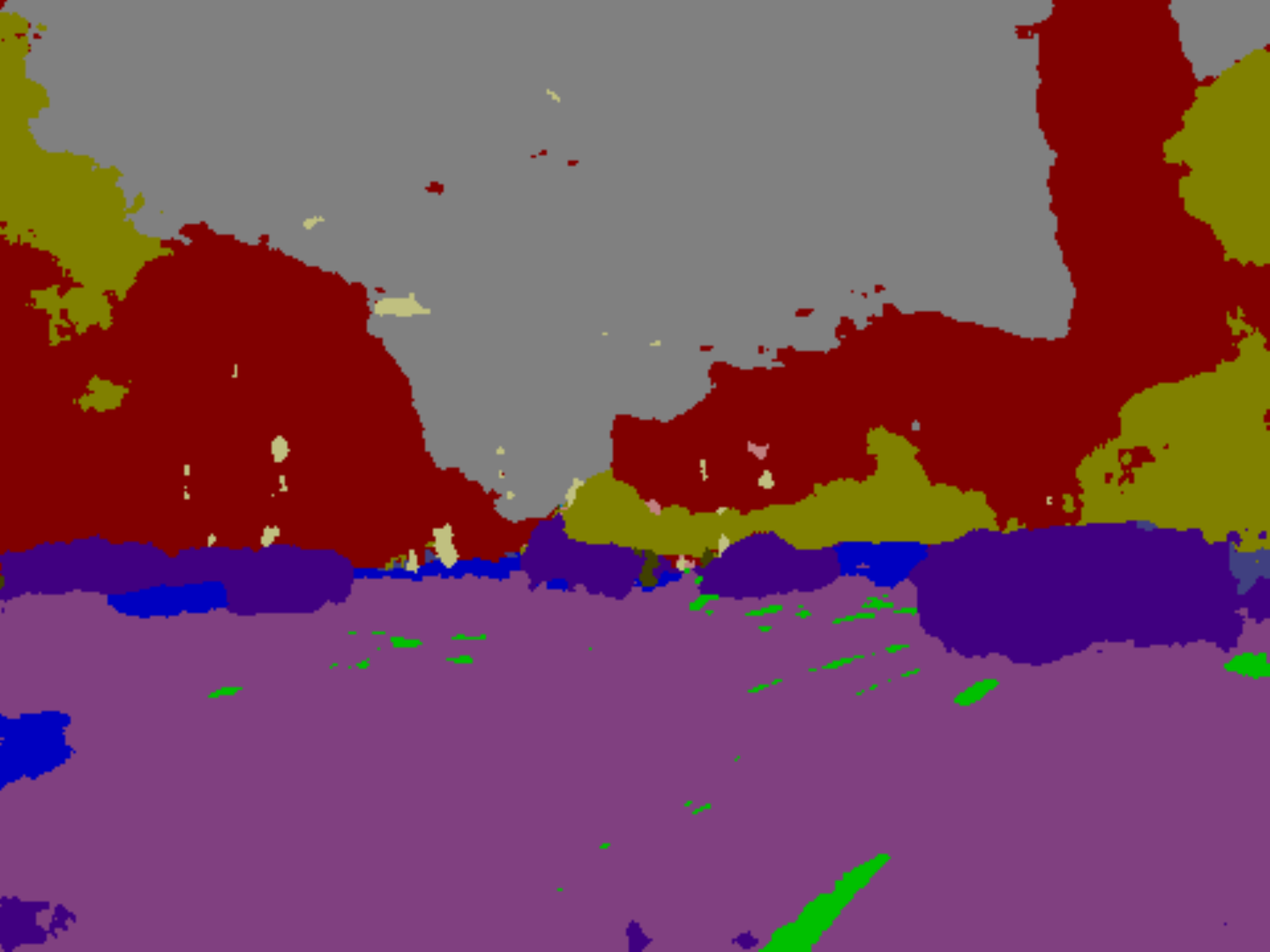}}
\label{fig:streetview_seg}
\vspace{-1.4em}
\caption*{StreetView}
\end{subfigure}\hspace*{\fill}
\centering
\caption{Sample images from the datasets used in the experiments: (top) RGB image, (middle) Depth image, (bottom) Semantic segmentation. (left) \emph{SYNTHIA} with perfect semantic segmentation, (middle left) \emph{SYNTHIA} with \emph{AdapNet} semantic segmentation, (middle right) \emph{Airsim} with perfect semantic segmentation, (right) \emph{StreetView} with \emph{SegNet} semantic segmentation.}
\vspace{-1.5em}
\label{fig:mosaic}
\end{figure}

Our approach relies on semantic representations of scenes.
While we do not propose contributions on semantic extraction from raw sensor data, recent advances on semantic segmentation show ever increasing accuracies on visual and depth data~\cite{noh2015learning, garcia2017review, badrinarayanan2015segnet, valada2017adapnet}.
We therefore evaluate the performance on \emph{SYNTHIA} both using semantic segmentation with \emph{AdapNet}~\cite{valada2017adapnet}, and the ground truth as provided by the dataset.
On the \emph{Airsim} data, we only use the segmentation from the dataset, and on the \emph{StreetView} dataset, we use semantic segmentation with \emph{SegNet}~\cite{badrinarayanan2015segnet}.
\subsection{Experimental Setup}
We evaluate the core components of \emph{X-View} in different experimental settings.
In all experiments, we evaluate \emph{X-View} on overlapping trajectories and the provided depth and segmentation images of the data.
First, we focus our evaluation of the different graph settings on the \emph{SYNTHIA} dataset.
We then perform a comparative analysis on \emph{SYNTHIA}, \emph{Airsim}, and \emph{StreetView}.

In \emph{SYNTHIA}, we use the left forward camera for building a database map and then use the left backward camera for localization.
Furthermore, we use 8 semantic classes of \emph{SYNTHIA}: \emph{building}, \emph{street}, \emph{sidewalk}, \emph{fence}, \emph{vegetation}, \emph{pole}, \emph{car}, and \emph{sign}, and reject the remaining four classes: \emph{sky}, \emph{pedestrian}, \emph{cyclist}, \emph{lanemarking}.
The \emph{AdapNet} semantic segmentation model is trained on other sequences of the \emph{SYNTHIA} dataset with different seasons and weather conditions.

Analogously, we use the forward-view of the car in the \emph{Airsim} dataset to build the database map and then localize the \ac{UAV} based on a downward-looking camera.
Here we use 6 classes \emph{(street, building, car, fence, hedge, tree)} and reject the remaining from insertion into the graph \emph{(powerline, pool, sign, wall, bench, rock)}, as these are usually only visible by one of the robots, or their scale is too small to be reliably detected from the aerial robot.

Finally, in the \emph{StreetView} data, we use the forward view to build the database and localize using the rear facing view.
Out of the 12 classes that we extract using the pre-trained \emph{SegNet} model\footnote{\href{goo.gl/EyReyn}{goo.gl/EyReyn}}, we use five, i.e., \emph{(road, sidewalk, vegetation, fence, car)}, and reject the remaining as these are either dynamic \emph{(pedestrian, cyclist)}, unreliably segmented \emph{(pole, road sign, road marking)}, or omni-present in the dataset \emph{(building, sky)}.

We build the graphs from consecutive frames in all experiments, and use the \emph{3D} information to connect and merge vertices and edges, as described in ~\ref{sec:extraction}.
The difference between graph construction in image- and \emph{3D}-space is evaluated in a separate experiment.
No assumptions are made on the prior alignment between the data.
The ground-truth alignment is solely used for performance evaluation.
\subsection{Localization performance}
\begin{figure*}[ht!]
\begin{subfigure}{0.33\textwidth}
\resizebox{\textwidth}{!}
{\includegraphics{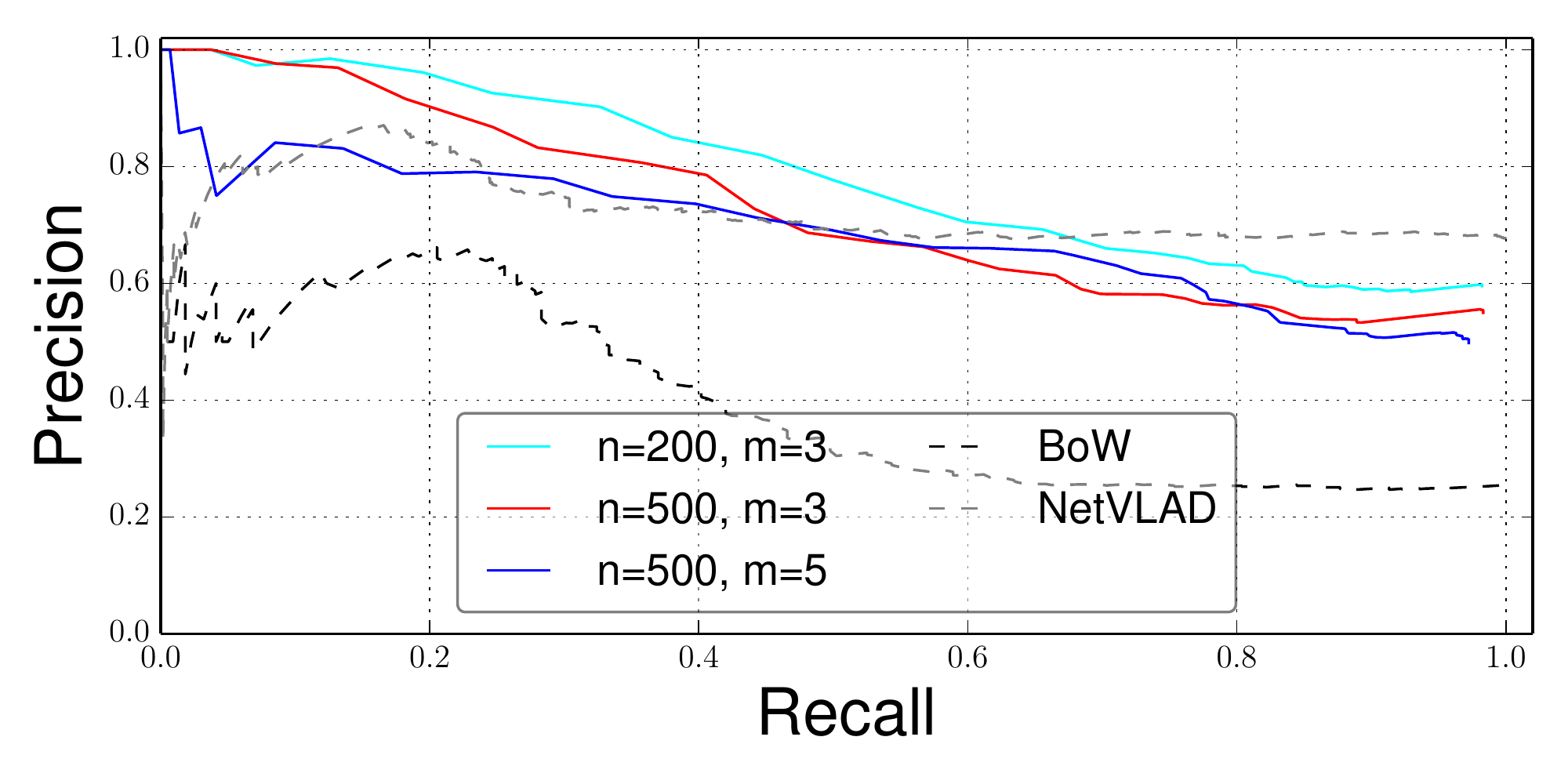}}
%{\input{figures/PR_walks.pgf}}
\vspace*{-7mm}
\caption{Descriptor parameters.}
\label{fig:pr_descriptors_synthia}
\end{subfigure}\hspace*{\fill}
\begin{subfigure}{0.33\textwidth}
\resizebox{\textwidth}{!}
{\includegraphics{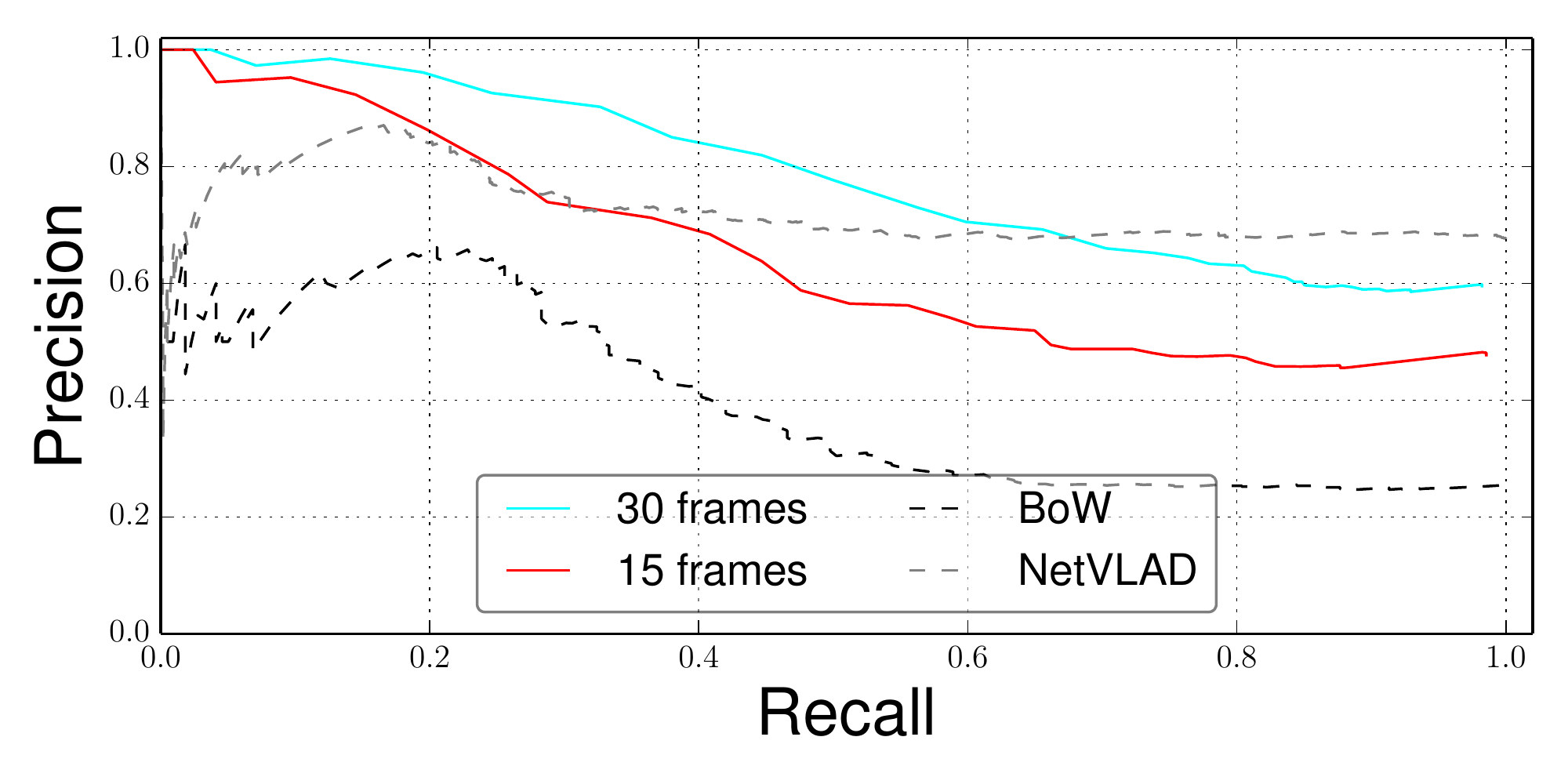}}
%{\input{figures/PR_query_length.pgf}}
\vspace*{-7mm}
\caption{Query length.}
\label{fig:pr_query_length_synthia}
\end{subfigure}\hspace*{\fill}
\begin{subfigure}{0.33\textwidth}
\resizebox{\textwidth}{!}
{\includegraphics{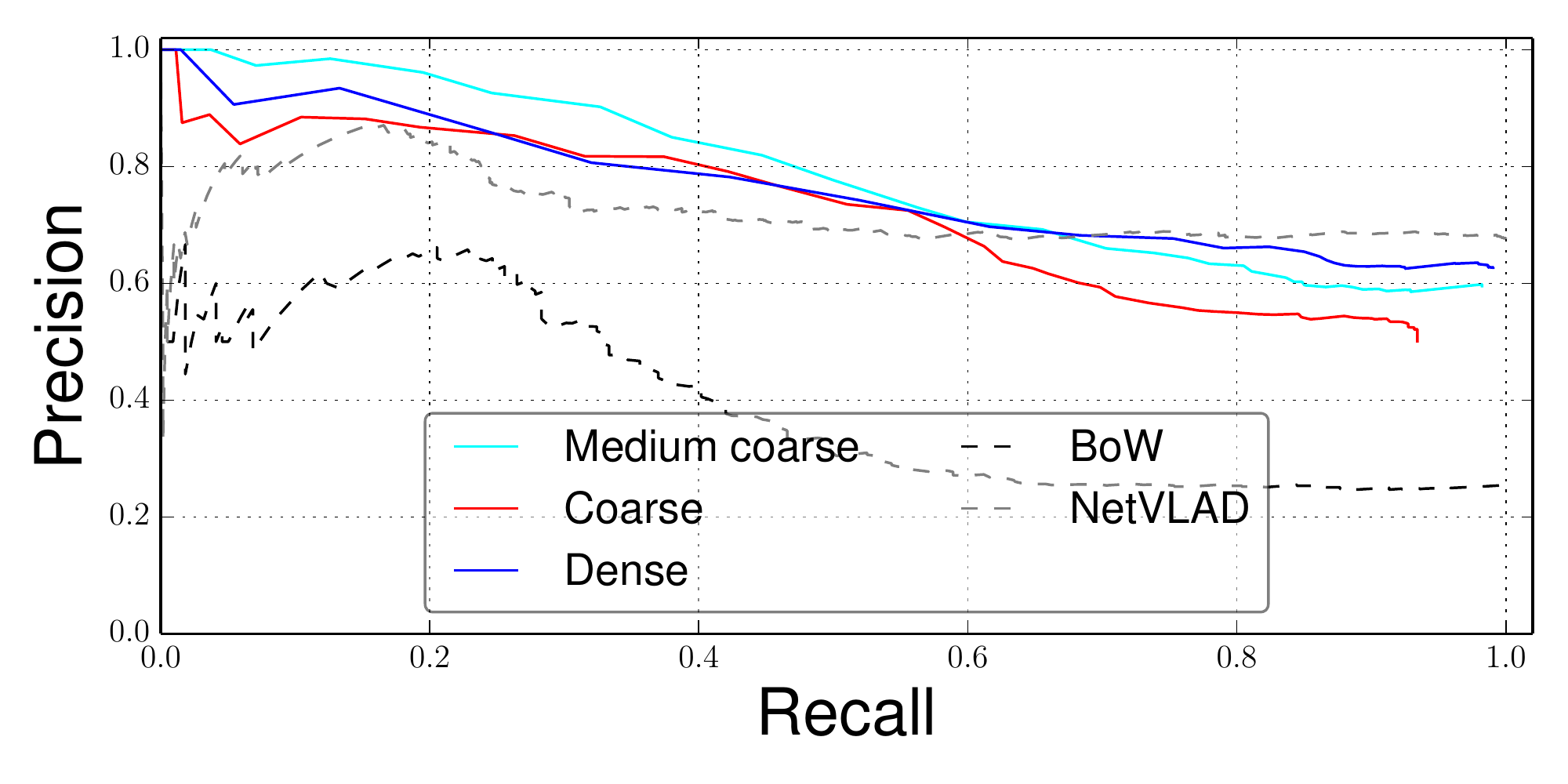}}
%{\input{figures/PR_coarseness.pgf}}
\vspace*{-7mm}
\caption{Graph coarseness.}
\label{fig:pr_coarseness_synthia}
\end{subfigure}\hspace*{\fill}
\vspace{-1.2mm}
\smallbreak
\begin{subfigure}{0.33\textwidth}
\resizebox{\textwidth}{!}
{\includegraphics{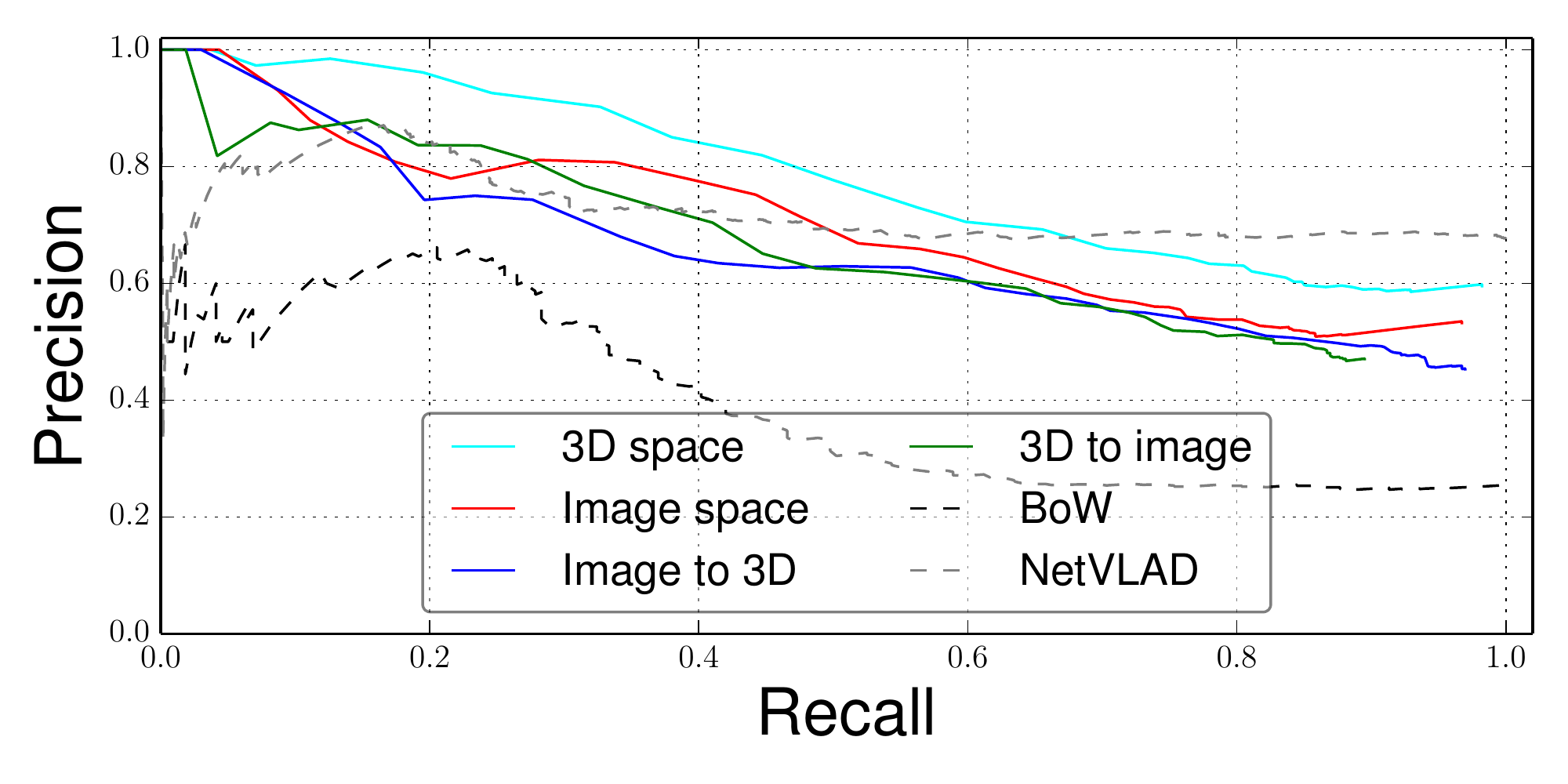}}
%{\input{figures/PR_image_space.pgf}}
\vspace*{-7mm}
\caption{Construction type.}
\label{fig:pr_imagespace_synthia}
\end{subfigure}\hspace*{\fill}
\begin{subfigure}{0.33\textwidth}
\resizebox{\textwidth}{!}
{\includegraphics{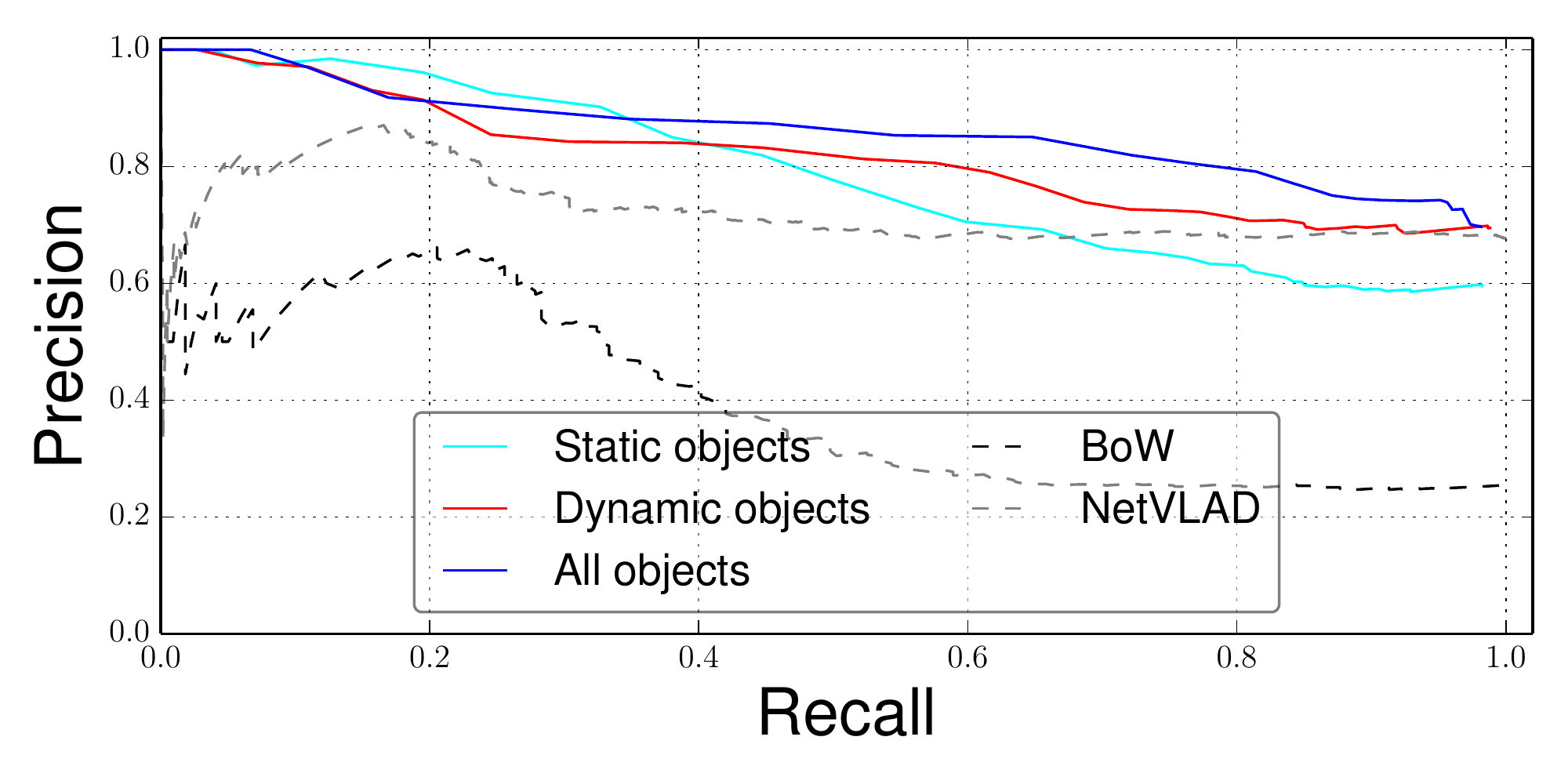}}
%{\input{figures/PR_dynamic_objects.pgf}}
\vspace*{-7mm}
\caption{Number of Semantic classes.}
\label{fig:pr_dynamic_objects}
\end{subfigure}\hspace*{\fill}
\begin{subfigure}{0.33\textwidth}
\resizebox{\textwidth}{!}{\includegraphics{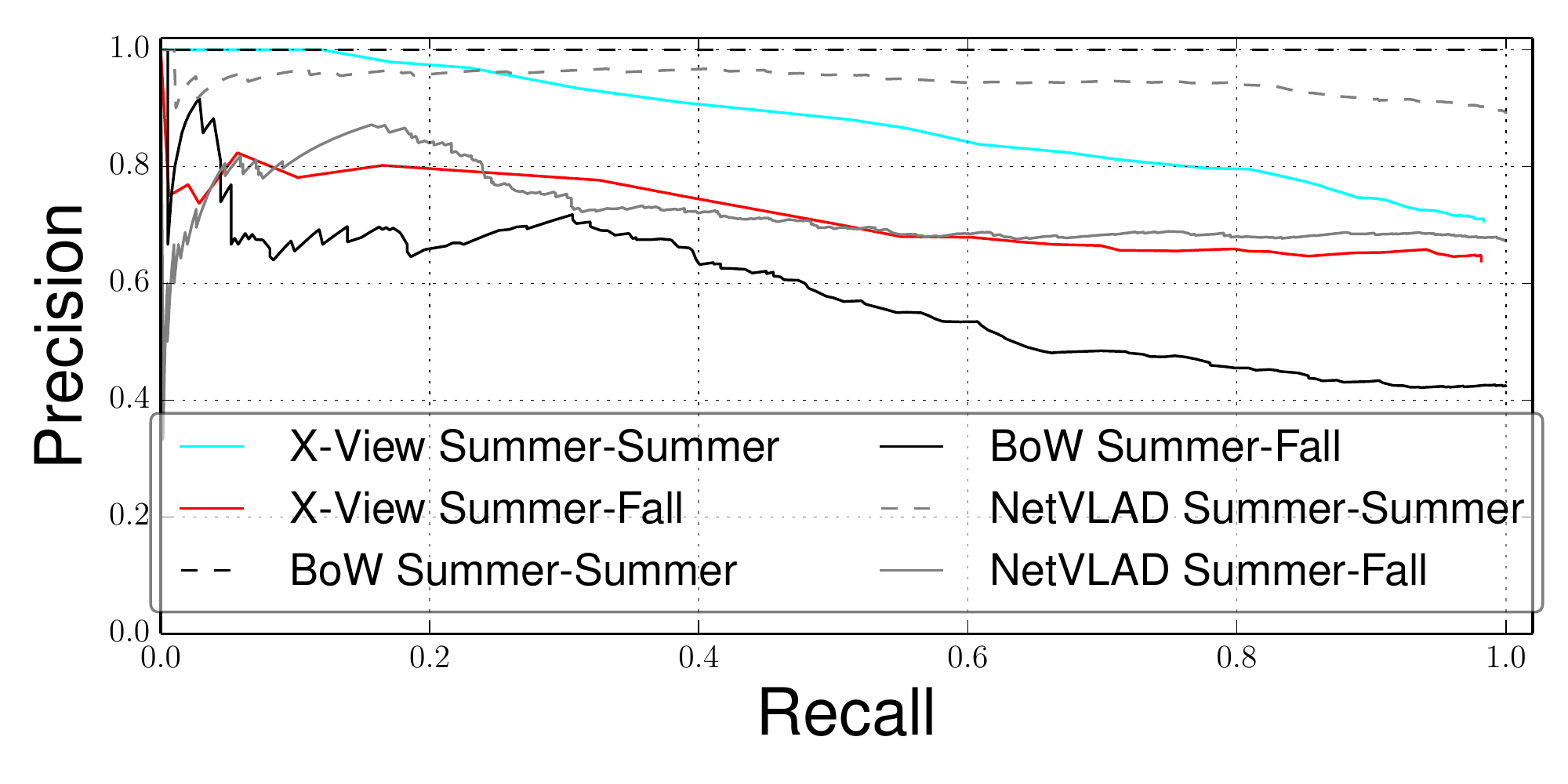}}
%\resizebox{\textwidth}{!}{\input{figures/PR_seasons.pgf}}
\vspace*{-7mm}
\caption{Seasonal changes on same view.}
\label{fig:pr_seasons}
\end{subfigure}\hspace*{\fill}
\caption{\ac{PR} curves for localization of the rear view semantic images against a database graph built from the forward view on the \emph{SYNTHIA} dataset (except (\subref{fig:pr_seasons})). 
For all plots we accept a localization if it falls within a distance of $20\,m$ from the ground-truth robot position. 
This threshold corresponds to the value up to which query graph vertices of the same semantic instance can be off from their corresponding location in the database graph, caused by the graph construction technique. 
(\subref{fig:pr_descriptors_synthia}) illustrates the effect of different descriptor settings on the localization performance. 
(\subref{fig:pr_query_length_synthia}) shows the effect of increasing the amount of frames used for query graph construction, while (\subref{fig:pr_coarseness_synthia}) depicts the effect of using coarser graphs, i.e., a large distance in which we merge vertices of same class label. 
In (\subref{fig:pr_imagespace_synthia}) we compare the extraction methods in image-, and \emph{3D}-space and in (\subref{fig:pr_dynamic_objects}) the effect of including all semantic objects against including a subset of semantic classes. 
Lastly, in (\subref{fig:pr_seasons}), we evaluate the localization performance on a configuration with the right frontal camera as query and the left frontal camera for the database, under the effect of seasonal changes.
In contrast to the other plots where we use the ground truth, we use semantic segmentation with \emph{AdapNet} on the data.
The appearance-based techniques used are visual \ac{BoW}~\cite{galvez2012bags} and NetVLAD~\cite{arandjelovic2016netvlad}.}
\label{fig:pr_plots}
\vspace{-5mm}
\end{figure*}
\begin{figure*}[ht!]
\centering
\begin{subfigure}{0.5\textwidth}
\resizebox{\textwidth}{!}
{\includegraphics{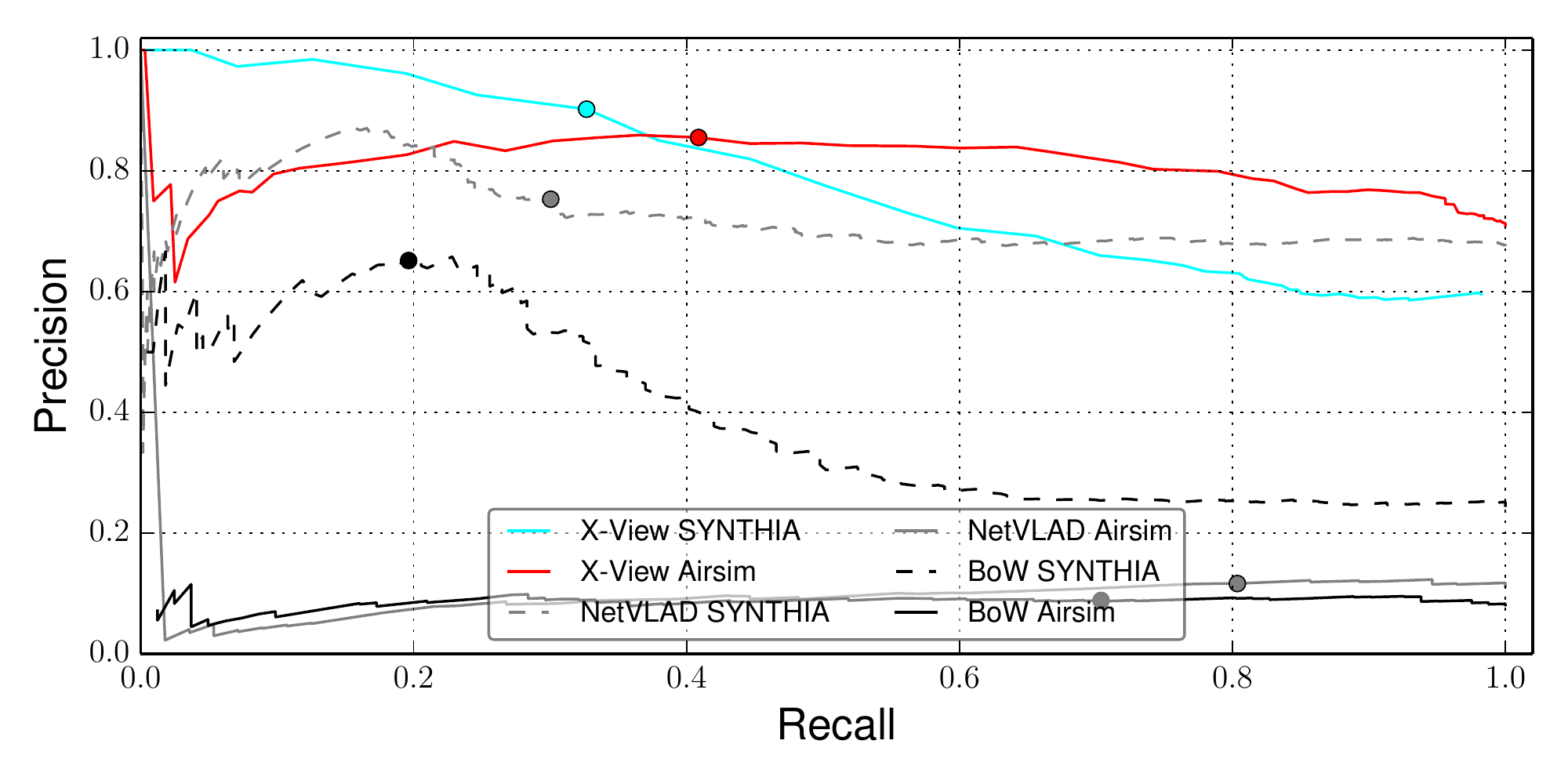}}
%{\input{figures/PR_datasets.pgf}}
\vspace{-7mm}
\caption{Perfect Semantic Segmentation.}
\label{fig:pr_datasets}
\end{subfigure}\hspace*{\fill}
\begin{subfigure}{0.5\textwidth}
\resizebox{\textwidth}{!}
{\includegraphics{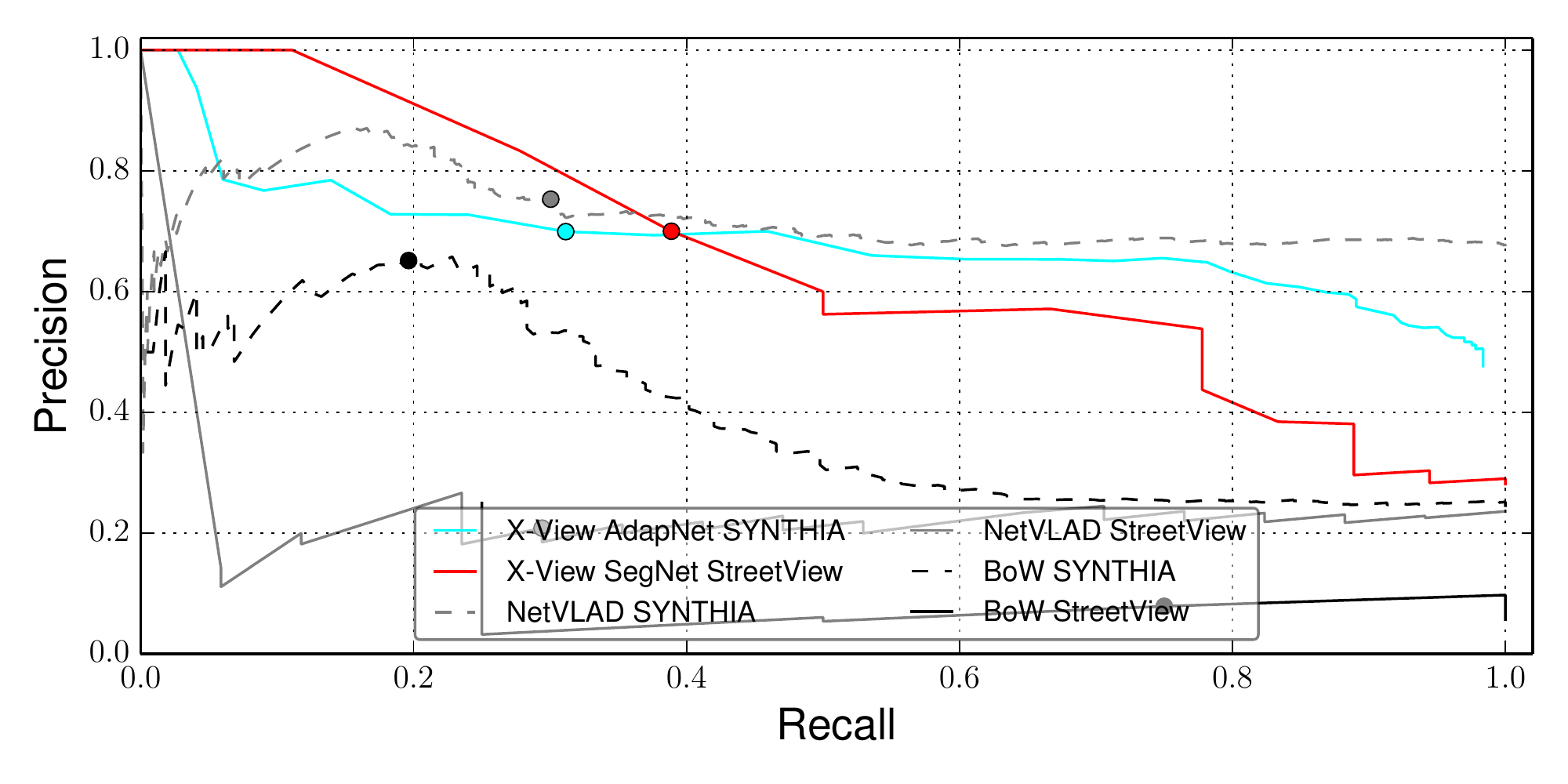}}
%{\input{figures/PR_datasets_real.pgf}}
\vspace{-7mm}
\caption{\ac{CNN}-based Semantic Segmentation.}
\label{fig:pr_datasets_real}
\end{subfigure}\hspace*{\fill}
%
%\vspace{-2mm}
\smallbreak
\begin{subfigure}{0.5\textwidth}
\resizebox{\textwidth}{!}
{\includegraphics{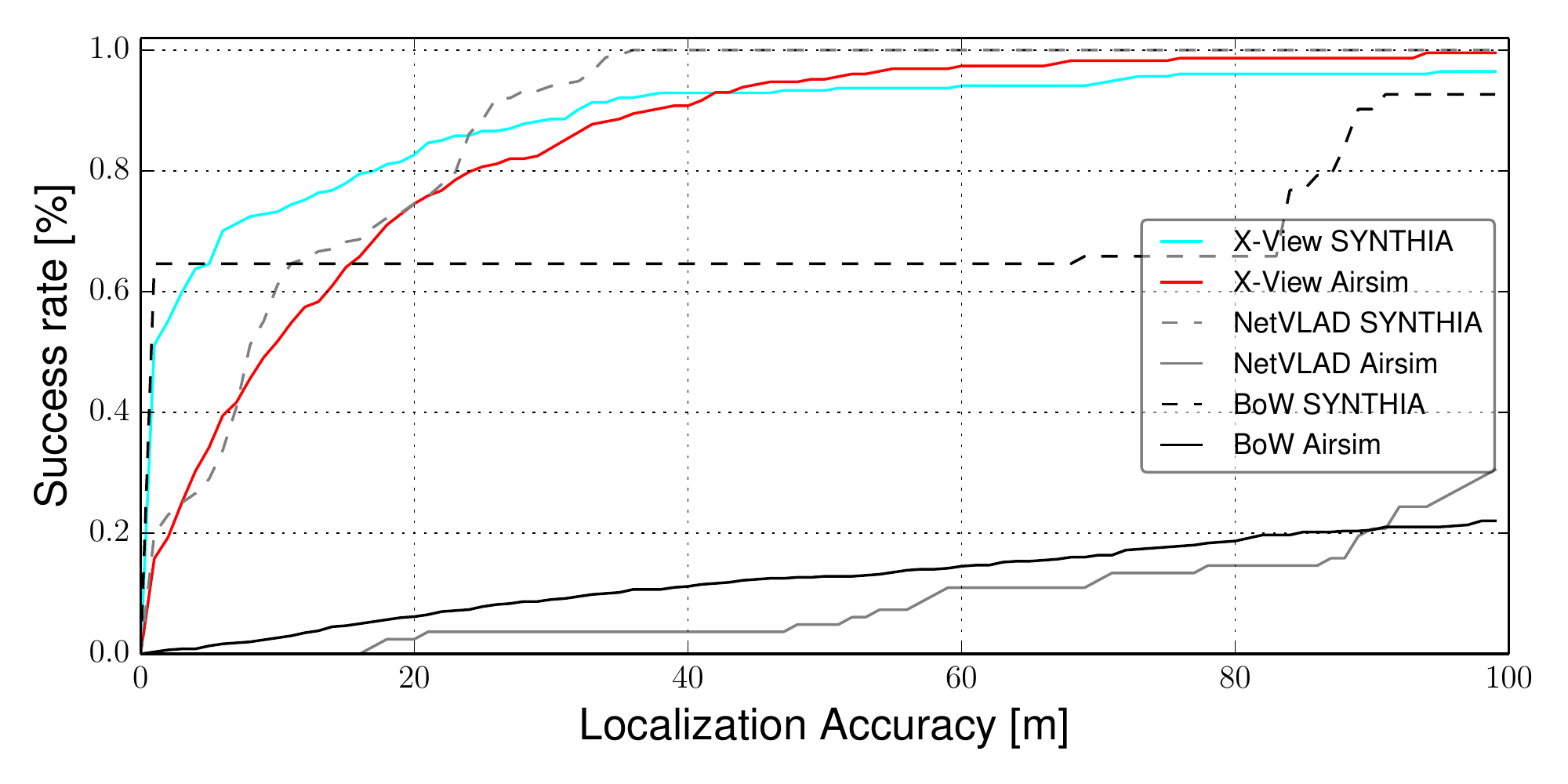}}
%{\input{figures/success_rate.pgf}}
\vspace{-7mm}
\caption{Perfect Semantic Segmentation.}
\label{fig:success_rate_perfect}
\end{subfigure}\hspace*{\fill}
\begin{subfigure}{0.5\textwidth}
\resizebox{\textwidth}{!}
{\includegraphics{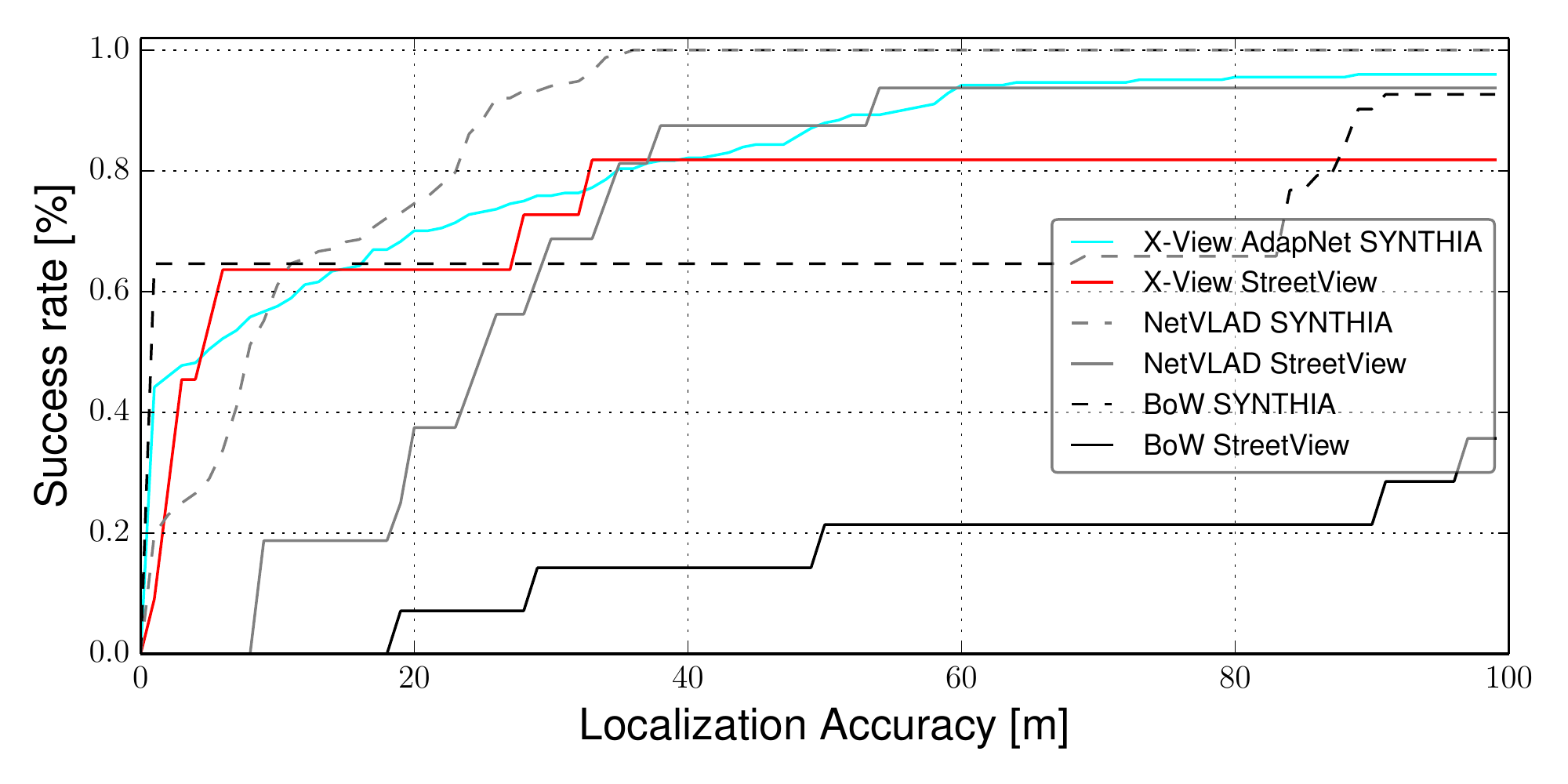}}
%{\input{figures/success_rate_datasets.pgf}}
\vspace{-7mm}
\caption{\ac{CNN}-based Semantic Segmentation.}
\label{fig:success_rate_segmentation}
\end{subfigure}\hspace*{\fill}
\caption{Localization performance of \emph{X-View} on the \emph{SYNTHIA}, \emph{Airsim}, and the \emph{StreetView} data compared to the appearance-based methods~\cite{galvez2012bags, arandjelovic2016netvlad}. 
The operation points are chosen according to the respective \ac{PR} curves in (\subref{fig:pr_datasets}) and (\subref{fig:pr_datasets_real}), indicated as dots.
(\subref{fig:success_rate_perfect}) illustrates the performance on perfectly semantically segmented data on \emph{SYNTHIA}, and \emph{Airsim}.  
(\subref{fig:success_rate_segmentation}) shows the system's performance on the \emph{SYNTHIA}, and \emph{StreetView} datasets using \ac{CNN}-based semantic segmentation.}
\vspace{-5mm}
\label{fig:success_rate}
\end{figure*}
We generate the \ac{PR} of the localization based on two thresholds.
The localization threshold $t_L$ is applied on the distance between the estimated robot position $\boldsymbol{c}_i^*$ and the ground truth position $\boldsymbol{c}_{gt}$.
It is set as \emph{true}, if the distance between $\boldsymbol{c}_i^*$ and $\boldsymbol{c}_{gt}$ is smaller than $t_L$, i.e., $\norm{\boldsymbol{c}_i^*-\boldsymbol{c}_{gt}} \leq t_L$, and to \emph{false} for $\norm{\boldsymbol{c}_i^*-\boldsymbol{c}_{gt}} > t_L$.
The margin $t_L$ on the locations is required, since $\boldsymbol{G}_q$ and $\boldsymbol{G}_{db}$ do not create vertices in the exact same spot.
The same node can be off by up to twice the distance that we use for merging vertices in a graph.
Here, we use $t_L=20\,m$ for \emph{SYNTHIA} and \emph{StreetView}, and $t_L=30\,m$ for \emph{Airsim}.
For the \ac{PR} curves, we vary the consistency threshold $t_c$ that is applied on the RANSAC-based rejection, i.e., the acceptable deviation from the consensus transformation between query and database graph vertices.
The localization estimation yields a positive vote for an estimated consensus value $s$ of $s \leq t_c$ and a negative vote otherwise.

Firstly, we evaluate the effect of different options on the description and matching using the random walk descriptors (i.e., random walk parameters, graph coarseness, number of query frames, dynamics classes, graph edge construction technique, and seasonal changes) as described in Sec.~\ref{sec:extraction}~-~\ref{sec:matching}.
To illustrate the contrast to appearance-based methods, we also present results on two visual place recognition techniques based on \ac{BoW}, as implemented by~\citet{galvez2012bags}, and NetVLAD~\cite{arandjelovic2016netvlad} on the datasets' RGB data.
To generate the \ac{PR} of the reference techniques, we vary a threshold on the inverse similarity score for \ac{BoW}, and a threshold on the matching residuals of NetVLAD.

Furthermore, we show the performance of the full global localization algorithm on the operating point taken from the \ac{PR} curves.
Our performance metric is defined as the percentage of correct localizations over the Euclidean distance between $\boldsymbol{c}_i^*$ and $\boldsymbol{c}_{gt}$.
As for \ac{BoW} and NetVLAD, we take localization as the best matching image.
The localization error is then computed as the Euclidean distance between associated positions of the matched image and the ground truth image.
To improve performance of the appearance-based methods, we select the operating points with high performances, i.e., high precisions in the \ac{PR} curves.
\subsection{Results}
While we illustrate the effects of different attributes of \emph{X-View} in Fig.~\ref{fig:pr_plots} as evaluated on \emph{SYNTHIA}, we then also show a comparison on all datasets in Fig.~\ref{fig:success_rate}.

Fig.~\ref{fig:pr_descriptors_synthia} depicts the effect of varying the random walk descriptors on the graph.
Here, a descriptor size with number of random walks $n=200$ and walk depth $m$ between $3-5$, depending on the size of $\boldsymbol{G}_q$ perform best.
Both decreasing $n$ or increasing $m$ leads to a decrease in performance.
These findings are expected, considering query graph sizes ranging between $20-40$ vertices.
Under these conditions, the graph can be well explored with the above settings.
Descriptors with larger walk depth $m$ significantly diverge between $\boldsymbol{G}_q$ and $\boldsymbol{G}_{db}$, as the random walk reaches the size limits of $\boldsymbol{G}_q$ and continues exploring already visited vertices, while it is possible to continue exploring $\boldsymbol{G}_{db}$ to greater depth.

Secondly, Fig.~\ref{fig:pr_query_length_synthia} presents \ac{PR}-curves for different sizes of $\boldsymbol{G}_q$, i.e., different numbers of frames used for the construction of $\boldsymbol{G}_q$.
An increase in the query graph size leads to a considerable increase of the localization performance.
Also this effect is expected as $\boldsymbol{G}_q$ contains more vertices, forming more unique descriptors.
However, it is also desirable to keep the size of $\boldsymbol{G}_q$ limited, as a growing query graph size requires larger overlap between $\boldsymbol{G}_q$ and $\boldsymbol{G}_{db}$.
Furthermore, the computational time for descriptor calculation and matching grows with increased query graph size.

Thirdly, Fig.~\ref{fig:pr_coarseness_synthia} shows the impact of increased graph coarseness, i.e., larger distances of merging vertices.
Here, the coarseness cannot be arbitrarily scaled to low or high values, as it leads to either over- or under-segmented graphs.
Our best performing results were obtained with a vertex merging distance of $10\,m$ for the \emph{SYNTHIA} dataset, and $15\,m$ for \emph{Airsim} and \emph{StreetView} datasets, respectively.

Fourthly, Fig.~\ref{fig:pr_imagespace_synthia} illustrates the effect of graph extraction in either image- or \emph{3D}-space.
The extraction in \emph{3D}-space, taking advantage of the depth information as described in Sec.~\ref{sec:extraction} shows superior performance.
However, \emph{X-View} still performs well when localizing a graph built in one space against a graph built in the other.

Fifthly, Fig.~\ref{fig:pr_dynamic_objects} explores the inclusion of different object classes.
The configurations are: Only static object classes, static object classes plus dynamic object classes, and all object classes.
Here, the results are not conclusive on the \emph{SYNTHIA} dataset and more evaluations will be needed in the future.

Lastly, Fig.~\ref{fig:pr_seasons} shows \emph{X-View}'s performance under seasonal change. 
We compare the performance of localizing the query graph built from the right forward facing camera of one season in the database graph built from the left forward facing camera of another season.
Here, we consider the summer and fall sequences of \emph{SYNTHIA}.
The \ac{BoW}-based techniques perform well in this scenario if the seasonal conditions are equal.
However, its performance drastically drops for inter-season localization, while \emph{X-View}, and NetVLAD suffer much less under the seasonal change.

The evaluation using \ac{PR}-curves, and success rates over the localization error is depicted in Fig.~\ref{fig:success_rate}.
\emph{X-View} has higher success rate in multi-view experiments than the appearance-based techniques on both synthetic datasets at our achievable accuracy of $20\,m$ for \emph{SYNTHIA} and $30\,m$ on \emph{Airsim} and using perfect semantic segmentation inputs as depicted in Fig.~\ref{fig:success_rate_perfect}.
These accuracies are considered successful as node locations between $\boldsymbol{G}_q$ and $\boldsymbol{G}_{db}$ can differ by twice the merging distance with our current graph merging strategy.
On the considered operation point of the \ac{PR} curve, \emph{X-View} achieves a localization accuracy of $85\,\%$ within $30\,m$ on \emph{Airsim}, and $85\,\%$ on \emph{SYNTHIA} within $20\,m$.

Furthermore, \emph{X-View} expresses comparable or better performance for multi-view localization than the appearance-based techniques using \ac{CNN}-based semantic segmentation on the \emph{SYNTHIA}, and \emph{StreetView} datasets respectively.
Here we consider successful localizations within $20\,m$ for both datasets.
The achieved accuracies on the chosen operation points are $70\,\%$ on \emph{SYNTHIA}, and $65\,\%$ on \emph{StreetView}.

Finally, we also report timings of the individual components of our system in Table~\ref{tab:times}.
Here, the construction of $\boldsymbol{G}_q$ has by far the largest contribution, due to iteratively matching and merging frames into $\boldsymbol{G}_q$.
As the graphs in \emph{SYNTHIA} consider more classes and smaller merging distances, these generally contain more vertices and therefore longer computational times.
\begin{table}[t]
\centering
\begin{tabular}{c|cc}
Module & \thead{\emph{SYNTHIA}\\ } & \thead{\emph{Airsim} \\}\\
\hline
\thead{Blob extraction} & \thead{$2.73 \pm 0.65$} & \thead{$1.76\pm 0.26$}\\
Construction of $\boldsymbol{G}_q$& \thead{$337.39\pm 92.81$} & \thead{$257.40\pm 28.30$}\\
\thead{Random Walks Generation} & \thead{$1.38 \pm 0.82$} & \thead{$1.07\pm 0.56$}\\
\thead{Matching $\boldsymbol{G}_q$ to $\boldsymbol{G}_{db}$} & \thead{$7.30\pm 4.51$} & \thead{$4.33\pm 1.25$}\\
\thead{Localization Back-End} & \thead{$22.50 \pm 9.71$} & \thead{$5.15 \pm 0.63$}\\
\hline
Total & $371.3 \pm 108.5 $ & $269.71\pm 31.0$ \\
\end{tabular}
\caption{Timing results in $ms$, reporting the means and standard deviations per frame on the best performing configurations on \emph{SYNTHIA} and \emph{Airsim}. 
The timings were computed on a single core of an Intel Xeon E3-1226 CPU @ 3.30GHz.}
\label{tab:times}
\vspace{-5mm}
\end{table}

\subsection{Discussion}
\label{sec:discussion}
Global registration of multi-view data is a difficult problem where traditional appearance based techniques fail.
Semantic graph representations can provide significantly better localization performance under these difficult perceptual conditions.
We furthermore give insights how different parameters, choices, and inputs' qualities affect the system's performance.
Our results obtained with \emph{X-View} show a better localization performance than appearance-based methods, such as \ac{BoW} and NetVLAD.

During our experiments, we observed that some of the parameters are dependent on each other.
Intuitively, the coarseness of the graph has an effect on the random walk descriptors as a coarser graph contains fewer vertices and therefore deeper random walks show decreasing performance as $\boldsymbol{G}_q$ can be explored with short random walks.
On the other hand, an increasing amount of frames used for localization has the reverse effect on the descriptor depth as $\boldsymbol{G}_q$ potentially contains more vertices, and deeper random walks do not show a performance drop as they do for smaller query graphs.

Also the success rate curves indicate that \emph{X-View} outperforms the appearance based methods particularly in the presence of strong view-point changes.
While the appearance-based methods fail to produce interesting results for the \emph{Airsim} dataset, they have a moderate to good amount of successful localizations on \emph{SYNTHIA} and \emph{StreetView}.
On the other hand, \emph{X-View} has generally higher localization performance and does not show a strong drop in performance among datasets.
While computational efficiency has not been the main focus of our research, the achieved timings are close to the typical requirements for robotic applications.

Finally, we performed experiments both using ground truth semantic segmentation inputs, and \ac{CNN}-based semantic segmentation. 
The performance with semantic segmentation using \emph{AdapNet}~\cite{valada2017adapnet} shows to be close to the achievable performance with ground truth segmentation on \emph{SYNTHIA}.
Using the \emph{SegNet}~\cite{badrinarayanan2015segnet} semantic segmentation on real image data from \emph{StreetView} demonstrates the effectiveness of our algorithm's full pipeline on real data, resulting in better performance than the best reference algorithm.
Despite the high performance, our system still receives a moderate amount of false localizations, which is due to similar sub-graphs at different locations, and we hope to mitigate this effect by including it into a full SLAM system in the future.

Furthermore, \emph{3D} locations of the vertices are presently positioned at the blob centers of their first observation.
We expect a more precise positioning technique to further disambiguate the associations between graphs.
\section{CONCLUSIONS}
\label{sec:conclusions}
In this paper we presented \emph{X-View}, a multi-view global localization algorithm leveraging semantic graph descriptor matching.
The approach was evaluated on one real-world and two simulated urban outdoor datasets with drastically different view-points.
Our results show the potential of using graph representations of semantics for large-scale robotic global localization tasks.
Alongside further advantages, such as compact representation and real-time-capability, the presented method is a step towards view-point invariant localization.

Our current research includes the investigation of more sophisticated graph construction methods, the integration of \emph{X-View} with a full SLAM system to generate loop closures, and learning-based class selection for discriminative representations.

\bibliographystyle{IEEEtranN}

\bibliography{eth}

% Generated by IEEEtranN.bst, version: 1.13 (2008/09/30)
\begin{thebibliography}{26}
\providecommand{\natexlab}[1]{#1}
\providecommand{\url}[1]{#1}
\csname url@samestyle\endcsname
\providecommand{\newblock}{\relax}
\providecommand{\bibinfo}[2]{#2}
\providecommand{\BIBentrySTDinterwordspacing}{\spaceskip=0pt\relax}
\providecommand{\BIBentryALTinterwordstretchfactor}{4}
\providecommand{\BIBentryALTinterwordspacing}{\spaceskip=\fontdimen2\font plus
\BIBentryALTinterwordstretchfactor\fontdimen3\font minus
  \fontdimen4\font\relax}
\providecommand{\BIBforeignlanguage}[2]{{%
\expandafter\ifx\csname l@#1\endcsname\relax
\typeout{** WARNING: IEEEtranN.bst: No hyphenation pattern has been}%
\typeout{** loaded for the language `#1'. Using the pattern for}%
\typeout{** the default language instead.}%
\else
\language=\csname l@#1\endcsname
\fi
#2}}
\providecommand{\BIBdecl}{\relax}
\BIBdecl

\bibitem[Cummins and Newman(2008)]{cummins2008fab}
M.~Cummins and P.~Newman, ``Fab-map: Probabilistic localization and mapping in
  the space of appearance,'' \emph{The International Journal of Robotics
  Research}, vol.~27, no.~6, pp. 647--665, 2008.

\bibitem[G{\'a}lvez-L{\'o}pez and Tardos(2012)]{galvez2012bags}
D.~G{\'a}lvez-L{\'o}pez and J.~D. Tardos, ``Bags of binary words for fast place
  recognition in image sequences,'' \emph{IEEE Transactions on Robotics},
  vol.~28, no.~5, pp. 1188--1197, 2012.

\bibitem[Lowry et~al.(2016)Lowry, S{\"u}nderhauf, Newman, Leonard, Cox, Corke,
  and Milford]{lowry2016visual}
S.~Lowry, N.~S{\"u}nderhauf, P.~Newman, J.~J. Leonard, D.~Cox, P.~Corke, and
  M.~J. Milford, ``Visual place recognition: A survey,'' \emph{IEEE
  Transactions on Robotics}, vol.~32, no.~1, pp. 1--19, 2016.

\bibitem[Arandjelovic et~al.(2016)Arandjelovic, Gronat, Torii, Pajdla, and
  Sivic]{arandjelovic2016netvlad}
R.~Arandjelovic, P.~Gronat, A.~Torii, T.~Pajdla, and J.~Sivic, ``Netvlad: Cnn
  architecture for weakly supervised place recognition,'' in \emph{IEEE
  Conference on Computer Vision and Pattern Recognition}, 2016, pp. 5297--5307.

\bibitem[Gawel et~al.(2016)Gawel, Cieslewski, Dub{\'e}, Bosse, Siegwart, and
  Nieto]{gawel2016structure}
A.~Gawel, T.~Cieslewski, R.~Dub{\'e}, M.~Bosse, R.~Siegwart, and J.~Nieto,
  ``Structure-based vision-laser matching,'' in \emph{IEEE/RSJ International
  Conference on Intelligent Robots and Systems (IROS)}, 2016, pp. 182--188.

\bibitem[Gawel et~al.(2017)Gawel, Dub{\'e}, Surmann, Nieto, Siegwart, and
  Cadena]{gawel20173d}
A.~Gawel, R.~Dub{\'e}, H.~Surmann, J.~Nieto, R.~Siegwart, and C.~Cadena, ``3d
  registration of aerial and ground robots for disaster response: An evaluation
  of features, descriptors, and transformation estimation,'' in \emph{IEEE
  International Symposium on Safety, Security}, 2017.

\bibitem[Chen et~al.(2017)Chen, Jacobson, Sunderhauf, Upcroft, Liu, Shen, Reid,
  and Milford]{chen2017deep}
Z.~Chen, A.~Jacobson, N.~Sunderhauf, B.~Upcroft, L.~Liu, C.~Shen, I.~Reid, and
  M.~Milford, ``Deep learning features at scale for visual place recognition,''
  2017.

\bibitem[Stumm et~al.(2016)Stumm, Mei, Lacroix, Nieto, Hutter, and
  Siegwart]{stumm2016robust}
E.~Stumm, C.~Mei, S.~Lacroix, J.~Nieto, M.~Hutter, and R.~Siegwart, ``Robust
  visual place recognition with graph kernels,'' in \emph{IEEE Conference on
  Computer Vision and Pattern Recognition}, 2016, pp. 4535--4544.

\bibitem[Su et~al.(2016)Su, Han, Harang, and Yan]{su2016fast}
Y.~Su, F.~Han, R.~E. Harang, and X.~Yan, ``A fast kernel for attributed
  graphs,'' in \emph{SIAM International Conference on Data Mining}, 2016, pp.
  486--494.

\bibitem[Garcia-Garcia et~al.(2017)Garcia-Garcia, Orts-Escolano, Oprea,
  Villena-Martinez, and Garcia-Rodriguez]{garcia2017review}
A.~Garcia-Garcia, S.~Orts-Escolano, S.~Oprea, V.~Villena-Martinez, and
  J.~Garcia-Rodriguez, ``A review on deep learning techniques applied to
  semantic segmentation,'' \emph{arXiv preprint arXiv:1704.06857}, 2017.

\bibitem[Valada et~al.(2017)Valada, Vertens, Dhall, and
  Burgard]{valada2017adapnet}
A.~Valada, J.~Vertens, A.~Dhall, and W.~Burgard, ``Adapnet: Adaptive semantic
  segmentation in adverse environmental conditions,'' in \emph{IEEE
  International Conference on Robotics and Automation (ICRA)}, 2017, pp.
  4644--4651.

\bibitem[Badrinarayanan et~al.(2017)Badrinarayanan, Kendall, and
  Cipolla]{badrinarayanan2015segnet}
V.~Badrinarayanan, A.~Kendall, and R.~Cipolla, ``Segnet: A deep convolutional
  encoder-decoder architecture for image segmentation,'' \emph{IEEE
  Transactions on Pattern Analysis and Machine Intelligence}, 2017.

\bibitem[Cook(1971)]{cook1971complexity}
S.~A. Cook, ``The complexity of theorem-proving procedures,'' in \emph{ACM
  symposium on Theory of computing}.\hskip 1em plus 0.5em minus 0.4em\relax
  ACM, 1971, pp. 151--158.

\bibitem[Milford and Wyeth(2012)]{milford2012seqslam}
M.~J. Milford and G.~F. Wyeth, ``Seqslam: Visual route-based navigation for
  sunny summer days and stormy winter nights,'' in \emph{IEEE International
  Conference on Robotics and Automation (ICRA)}, 2012, pp. 1643--1649.

\bibitem[Cieslewski et~al.(2016)Cieslewski, Stumm, Gawel, Bosse, Lynen, and
  Siegwart]{cieslewski2016point}
T.~Cieslewski, E.~Stumm, A.~Gawel, M.~Bosse, S.~Lynen, and R.~Siegwart, ``Point
  cloud descriptors for place recognition using sparse visual information,'' in
  \emph{IEEE International Conference on Robotics and Automation (ICRA)}, 2016,
  pp. 4830--4836.

\bibitem[B{\"u}rki et~al.(2016)B{\"u}rki, Gilitschenski, Stumm, Siegwart, and
  Nieto]{burki2016appearance}
M.~B{\"u}rki, I.~Gilitschenski, E.~Stumm, R.~Siegwart, and J.~Nieto,
  ``Appearance-based landmark selection for efficient long-term visual
  localization,'' in \emph{IEEE International Conference on Intelligent Robots
  and Systems (IROS)}, 2016, pp. 4137--4143.

\bibitem[Sunderhauf et~al.(2015)Sunderhauf, Shirazi, Jacobson, Dayoub,
  Pepperell, Upcroft, and Milford]{sunderhauf2015place}
N.~Sunderhauf, S.~Shirazi, A.~Jacobson, F.~Dayoub, E.~Pepperell, B.~Upcroft,
  and M.~Milford, ``Place recognition with convnet landmarks: Viewpoint-robust,
  condition-robust, training-free,'' \emph{Robotics: Science and Systems},
  2015.

\bibitem[Huang and Beevers(2005)]{huang2005topological}
W.~H. Huang and K.~R. Beevers, ``Topological map merging,'' \emph{The
  International Journal of Robotics Research}, vol.~24, no.~8, pp. 601--613,
  2005.

\bibitem[Marinakis and Dudek(2010)]{marinakis2010pure}
D.~Marinakis and G.~Dudek, ``Pure topological mapping in mobile robotics,''
  \emph{IEEE Transactions on Robotics}, vol.~26, no.~6, pp. 1051--1064, 2010.

\bibitem[Kostavelis and Gasteratos(2015)]{kostavelis2015semantic}
I.~Kostavelis and A.~Gasteratos, ``Semantic mapping for mobile robotics tasks:
  A survey,'' \emph{Robotics and Autonomous Systems}, vol.~66, pp. 86--103,
  2015.

\bibitem[Bowman et~al.(2017)Bowman, Atanasov, Daniilidis, and
  Pappas]{bowman2017probabilistic}
S.~L. Bowman, N.~Atanasov, K.~Daniilidis, and G.~J. Pappas, ``Probabilistic
  data association for semantic slam,'' in \emph{IEEE International Conference
  on Robotics and Automation (ICRA)}, 2017, pp. 1722--1729.

\bibitem[Atanasov et~al.(2014)Atanasov, Zhu, Daniilidis, and
  Pappas]{atanasov2014semantic}
N.~Atanasov, M.~Zhu, K.~Daniilidis, and G.~J. Pappas, ``Semantic localization
  via the matrix permanent.'' in \emph{Robotics: Science and Systems}, 2014.

\bibitem[Perozzi et~al.(2014)Perozzi, Al-Rfou, and Skiena]{perozzi2014deepwalk}
B.~Perozzi, R.~Al-Rfou, and S.~Skiena, ``Deepwalk: Online learning of social
  representations,'' 2014, pp. 701--710.

\bibitem[Ros et~al.(2016)Ros, Sellart, Materzynska, Vazquez, and
  Lopez]{ros2016synthia}
G.~Ros, L.~Sellart, J.~Materzynska, D.~Vazquez, and A.~M. Lopez, ``The synthia
  dataset: A large collection of synthetic images for semantic segmentation of
  urban scenes,'' in \emph{IEEE Conference on Computer Vision and Pattern
  Recognition}, 2016, pp. 3234--3243.

\bibitem[Shah et~al.(2017)Shah, Dey, Lovett, and Kapoor]{airsim2017fsr}
S.~Shah, D.~Dey, C.~Lovett, and A.~Kapoor, ``Airsim: High-fidelity visual and
  physical simulation for autonomous vehicles,'' in \emph{Field and Service
  Robotics}, 2017.

\bibitem[Noh et~al.(2015)Noh, Hong, and Han]{noh2015learning}
H.~Noh, S.~Hong, and B.~Han, ``Learning deconvolution network for semantic
  segmentation,'' in \emph{IEEE International Conference on Computer Vision},
  2015, pp. 1520--1528.

\end{thebibliography}

\begin{acronym}
\acro{MAP}{Maximum a Posteriori}
\acro{UAV}{Unmanned Aerial Vehicle}
\acro{PR}{Precision-Recall}
\acro{BoW}{Bag of Words}
\acro{KNN}{K Nearest Neighbors}
\acro{EM}{Expectation Maximization}
\acro{CNN}{Convolutional Neural Network}
\end{acronym}

\end{document}